%% file: main.tex
\documentclass{article}

\usepackage{url}            
\usepackage[utf8]{inputenc} 
\usepackage[T1]{fontenc}    
\usepackage{booktabs}       
\usepackage{amsfonts}       
\usepackage{nicefrac}       
\usepackage{microtype}      

\usepackage{listings}       
\usepackage{graphicx}
\usepackage{caption}
\usepackage{subcaption}
\usepackage{amsmath}
\usepackage{amssymb}
\usepackage[font=small]{caption}
\usepackage{mathtools}
\usepackage{dsfont}
\usepackage{multirow}
\usepackage{tabularx}
\usepackage{arydshln}
\usepackage{pifont}
\usepackage{algorithm}
\usepackage{algorithmic}
\usepackage{courier}
\usepackage{bbm }
\usepackage{wrapfig}

\usepackage[nottoc]{tocbibind}
\usepackage{minitoc}

\usepackage{enumitem}
\setlist[itemize]{leftmargin=5mm, nolistsep}

\usepackage{color}
\usepackage[linewidth=0.5pt]{mdframed}

\usepackage{microtype}
\usepackage{graphicx}
\usepackage{booktabs} 

\usepackage[colorlinks=True, citecolor=blue, linkcolor=red]{hyperref}       

\usepackage[accepted]{icml2023}

\usepackage{amsthm}

\usepackage{titlesec}
\titleformat{\subsubsection}[hang]{\bfseries}{\thesubsubsection}{1em}{}

\usepackage[capitalize,noabbrev]{cleveref}

\theoremstyle{plain}

\theoremstyle{definition}

\theoremstyle{remark}

\usepackage[textsize=tiny]{todonotes}

\icmltitlerunning{Hierarchical Programmatic Reinforcement Learning}

\input{macro}

\begin{document}

\twocolumn[
\icmltitle{Hierarchical Programmatic Reinforcement Learning \break via Learning to Compose Programs}



\icmlsetsymbol{equal}{*}

\begin{icmlauthorlist}
\icmlauthor{Guan-Ting Liu}{equal,ntu}
\icmlauthor{En-Pei Hu}{equal,ntu}
\icmlauthor{Pu-Jen Cheng}{ntu}
\icmlauthor{Hung-Yi Lee}{ntu}
\icmlauthor{Shao-Hua Sun}{ntu}
\end{icmlauthorlist}

\icmlaffiliation{ntu}{National Taiwan University, Taipei, Taiwan}

\icmlcorrespondingauthor{Shao-Hua Sun}{shaohuas@ntu.edu.tw}

\icmlkeywords{Machine Learning, ICML}

\vskip 0.3in
]



\printAffiliationsAndNotice{\icmlEqualContribution} 

\begin{abstract}
\input{text/0_abstract}
\end{abstract}

\input{text/1_introduction}
\input{text/2_related_work}
\input{text/3_problem}
\input{text/4_approach}
\input{text/5_experiment}
\input{text/6_conclusion}
\input{text/7_ack}

\clearpage

\bibliography{ref}
\bibliographystyle{icml2023}

\newpage
\appendix
\onecolumn
\input{text/appendix}

\end{document}

%% file: macro.tex
\newcommand{\myshorttitle}{Hierarchical Programmatic Reinforcement Learning}

\newcommand{\sun}[1]{{\color{blue}{\small\bf\sf [Sun: #1]}}}
\newcommand{\guanting}[1]{{\color{orange}{\small\bf\sf [GT: #1]}}}
\newcommand{\enpei}[1]{{\color{green}{\small\bf\sf [EP: #1]}}}
\newcommand{\hungyi}[1]{{\color{cyan}{\small\bf\sf [HY: #1]}}}
\newcommand{\Skip}[1]{}

\newcommand{\method}[1]{HPRL}

\newcommand{\ie}{\textit{i}.\textit{e}.,\ }
\newcommand{\eg}{\textit{e}.\textit{g}.,\ }

\newcommand{\myfig}[1]{Figure \ref{#1}}
\newcommand{\mytable}[1]{Table \ref{#1}}
\newcommand{\myeq}[1]{Eq. \ref{#1}}
\newcommand{\mysecref}[1]{Section \ref{#1}}
\newcommand{\myalgo}[1]{Algorithm \ref{#1}}

\newcommand{\dotieconcat}[2]{
  \text{\raisebox{.8ex}{$\smallfrown$}}%
}

\makeatletter
\newcommand\dslfontsize{\@setfontsize\dslfontsize\@viiipt\@viiiipt}
\makeatother

\newcommand{\myparagraph}[1]{\noindent \textbf{#1.}}
\newcommand{\vspacesection}[1]{\vspace{-0.0cm}
\section{#1}
\vspace{-0.0cm}}
\newcommand{\vspacesubsection}[1]{\vspace{-0.0cm}
\subsection{#1}
\vspace{-0.0cm}}
\newcommand{\vspacesubsubsection}[1]{\vspace{-0.0cm}
\subsubsection{#1}
\vspace{-0.0cm}}

\newcommand\blfootnote[1]{%
  \begingroup
  \renewcommand\thefootnote{}\footnote{#1}%
  \addtocounter{footnote}{-1}%
  \endgroup
}

\newcommand\SmallCaption[1]{%
  \captionsetup{font=normalsize}%
  \caption{#1}}

\usepackage{color}
\usepackage{xcolor}
\definecolor{codegray}{rgb}{0.5,0.5,0.5}

\lstdefinelanguage{DSL}{
  sensitive = true,
  keywords={DEF, WHILE, IF},
  otherkeywords={
    >, <, ==
  },
  keywords = [2]{noMarkersPresent, leftIsClear, rightIsClear, frontIsClear, markersPresent},
  keywords = [3]{move, turnLeft, turnRight, pickMarker, putMarker},
  numbersep=8pt,
  tabsize=1,
  showstringspaces=false,
  breaklines=true,
  frame=top,
  comment=[l]{//},
  morecomment=[s]{/*}{*/},
  commentstyle=\color{purple}\ttfamily,
  stringstyle=\color{red}\ttfamily,
  belowskip=0.5pt,
  aboveskip=0.2pt,
  morestring=[b]',
  morestring=[b]",
  keywordstyle=\color{blue},
  keywordstyle=[2]\color{red},
  keywordstyle=[3]\color{codegray},
  numberstyle=\tiny\color{codegray},
  basicstyle=\ttfamily\tiny,
  }
\lstloadaspects{formats}
\lstdefinestyle{numbers}
{numbers=left, stepnumber=1, numberstyle=\tiny, numbersep=10pt, numberstyle=\tiny\color{codegray}}

%% file: text/0_abstract.tex
Aiming to produce reinforcement learning (RL) policies that are human-interpretable and can generalize better to novel scenarios, \citet{trivedi2021learning} present a method (LEAPS) that first learns a program embedding space to continuously parameterize diverse programs from a pre-generated program dataset, and then searches for a task-solving program in the learned program embedding space when given a task. Despite the encouraging results, the program policies that LEAPS can produce are limited by the distribution of the program dataset. Furthermore, during searching, LEAPS evaluates each candidate program solely based on its return, failing to precisely reward correct parts of programs and penalize incorrect parts. To address these issues, we propose to learn a meta-policy that composes a series of programs sampled from the learned program embedding space. By learning to compose programs, our proposed hierarchical programmatic reinforcement learning (HPRL) framework can produce program policies that describe out-of-distributionally complex behaviors and directly assign credits to programs that induce desired behaviors. The experimental results in the Karel domain show that our proposed framework outperforms baselines. The 
ablation studies confirm the limitations of LEAPS and justify our design choices.


%% file: text/1_introduction.tex
\vspacesection{Introduction}
\label{sec:intro}

Deep reinforcement learning (DRL) leverages the recent advancement in deep learning by reformulating the reinforcement learning problem as learning policies or value functions parameterized by deep neural networks. DRL has achieved tremendous success in various domains, including controlling robots~\cite{gu2017deep, ibarz2021train, lee2019composing, lee2021generalizable}, playing board games~\cite{silver2016mastering, silver2017mastering}, and strategy games~\cite{vinyals2019grandmaster, wurman2022outracing}. 
Yet, the black-box nature of neural network-based policies makes it difficult for the DRL-based systems to be interpreted and therefore trusted by human users~\cite{lipton2018mythos, PuiuttaV20}. 
Moreover, policies learned by DRL methods tend to overfit and often fail to generalize~\cite{zhang2018study, cobbe2019quantifying, sun2020program, liu2022improving}.\blfootnote{Project page: \url{https://nturobotlearninglab.github.io/hprl}}

To address the abovementioned issues of DRL, 
programmatic RL methods~\cite{bastani2018verifiable, Inala2020Synthesizing, landajuela21a, verma2018programmatically} explore various of more structured representations of policies, such as decision trees and state machines.
In particular, \citet{trivedi2021learning} present a framework, \textbf{L}earning \textbf{E}mbeddings for l\textbf{A}tent \textbf{P}rogram \textbf{S}ynthesis (LEAPS), that is designed to produce more interpretable and generalizable policies. Specifically, it aims to produce program policies structured in a given domain-specific language (DSL), which can be executed to yield desired behaviors. To this end, LEAPS first learns a program embedding space to continuously parameterize diverse programs from a pre-generated program dataset, and then searches for a task-solving program in the learned program embedding space when given a task described by a Markov Decision Process (MDP). The program policies produced by LEAPS are not only human-readable but also achieve competitive performance and demonstrate superior generalization ability.

Despite its encouraging results, LEAPS has two fundamental limitations.
\textit{Limited program distribution}: the program policies that LEAPS can produce are limited by the distribution of the pre-generated program dataset used for learning the program embedding space. 
This is because LEAPS is designed to search for a task-solving program from the learned embedding space, 
which inherently assumes that such a program is within the distribution of the program dataset. 
Such design makes it difficult for LEAPS to synthesize programs that are out-of-distributionally long or complex.
\textit{Poor credit assignment}: during the search for the task-solving program embedding,
LEAPS evaluates each candidate program solely based on the
cumulative discounted return of the program execution trace.
Such design fails to accurately attribute rewards obtained during the execution trajectories
to corresponding parts in synthesized programs 
or penalize program parts that induce incorrect behaviors.

This work aims to address the issues of limited program distribution and poor credit assignment. To this end, we propose a hierarchical programmatic reinforcement learning (HPRL) framework. Instead of searching for a program from a  program embedding space, we propose to learn a meta-policy, whose action space is the learned program embedding space, to produce a series of programs (\ie predict a sequence of actions) to yield a composed task-solving program. By re-formulating synthesizing a program as predicting a sequence of programs, HPRL can produce out-of-distributionally long or complex programs. Also, rewards obtained from the environment by executing each program from the composed program can be accurately attributed to each program, leading to more efficient learning.

To evaluate our proposed method, we adopt the Karel domain~\cite{pattis1981karel}, which features an agent that can navigate a grid world and interact with objects. Our method outperforms all the baselines by large margins on a problem set proposed in~\cite{trivedi2021learning}. To investigate the limitation of our method, we design a more challenging problem set on which our method consistently achieves better performance compared to LEAPS. Moreover, we inspect LEAPS' issues of limited program distribution and poor credit assignment with two experiments and demonstrate that our proposed method addresses these issues. We present a series of ablation studies to justify our design choices, including the reinforcement learning algorithms used to learn the meta-policy, and the dimensionality of the program embedding space.

%% file: text/2_related_work.tex
\vspacesection{Related Work}
\label{sec:related_work}

\myparagraph{Program Synthesis} 
Program synthesis methods concern automatically
synthesize programs that can transform some inputs to
desired outputs.
Encouraging results have been achieved in a variety of domains, including
string transformation~\citep{devlin2017robustfill, hong2020latent, zhong2023hierarchical}, 
array/tensor transformation~\citep{balog2016deepcoder, ellis2020dreamcoder},
computer commands~\citep{lin2018nl2bash, chen2021openaicodex, li2022alphacode}, 
graphics and 3D shape programs~\citep{nsd, liu2018learning, tian2018learning}, 
and describing behaviors of agents~\citep{bunel2018leveraging, sun2018neural, shin2018improving, chen2019executionguided, liao2019synthesizing, silver2020few}.
Most existing program synthesis methods consider
task specifications such as input/output pairs, demonstrations, or natural language descriptions.
In contrast, we aim to synthesize programs as policies
that can be executed to induce behaviors 
which maximize rewards defined by reinforcement learning tasks.

\myparagraph{Programmatic Reinforcement Learning} 
Programmatic reinforcement learning methods~\citep{choi2005learning, distill, sun2020program} explore 
various programmatic and more structured representations of policies,
including decision trees~\citep{bastani2018verifiable}, 
state machines~\cite{Inala2020Synthesizing},
symbolic expressions~\citep{landajuela21a},
and programs drawn from a domain-specific language~\cite{silver2020few, verma2018programmatically, verma2019imitation}.
Our work builds upon~\cite{trivedi2021learning}, whose goal is 
to produce program policies from rewards.
We aim to address the fundamental limitations of this work
by learning to compose programs to yield more expressive programs.

\myparagraph{Hierarchical Reinforcement Learning (HRL)} 
HRL frameworks
~\cite{sutton1999between, barto2003recent, vezhnevets2017feudal, bacon2017option}
aims to learn to operate on 
different levels of temporal abstraction,
allowing for learning or exploring more efficiently in sparse-reward environments.
In this work, instead of operating on pre-defined or learned temporal abstraction,
we are interested in learning with a level of abstraction defined by
a learned program embedding space to hierarchically compose programs.
One can view a learned program embedding space as 
continuously parameterized options or low-level policies. 

\myparagraph{Program Induction}
Program induction~\cite{graves2014neural, zaremba2015reinforcement, reed2016neural, dong2018neural, ellis2021dreamcoder} aims to perform tasks by inducing latent programs for improved generalization.
Contrary to our work, these methods neither explicitly synthesize programs nor address RL tasks.

%% file: text/3_problem.tex

\vspacesection{Problem Formulation}
\label{sec:problem}

\input{figure/dsl.tex}

Our goal is to develop a method that can synthesize a domain-specific, task-solving program 
which can be executed to interact with an environment and
maximize a discounted return defined by a Markov Decision Process.

\noindent \textbf{Domain Specific Language.} 
In this work, we adapt the domain-specific language (DSL) for the Karel domain used in~\cite{bunel2018leveraging, chen2019executionguided, trivedi2021learning}, shown in~\myfig{fig:dsl}.
This DSL is designed to describe the behaviors of the Karel agent, consisting of control flows, agent's perceptions, and agent's actions.
Control flows such as \texttt{if}, \texttt{else}, and \texttt{while} are allowed for 
describing diverging or repetitive behaviors.
Furthermore, Boolean and logical operators such as 
\texttt{and}, \texttt{or}, and \texttt{not} can be included to 
express more sophisticated conditions.
Perceptions such as 
\texttt{frontIsClear} and \texttt{markerPresent}
are defined based on situations in an environment which
can be perceived by an agent.
On the other hand, actions such as
\texttt{move}, \texttt{turnRight}, and \texttt{putMarker}, 
describe the primitive behaviors that an agent can perform 
in an environment.
A program policy considered in our work is structured in this DSL and can be executed to produce a sequence of actions based on perceptions.

\noindent \textbf{Markov Decision Process (MDP).} 
The tasks considered in this work are defined by 
finite-horizon discounted MDPs.
The performance of a policy with its rollout 
(a sequence of states and actions $\{(s_0 , a_0), ... , (s_t , a_t)\}$) 
is evaluated based on a discounted return $\sum_{t=0}^{T} \gamma^t r_t$, 
where $r_t = \mathcal{R}(s_t, a_t)$ indicates the reward function and
$T$ is the horizon of the episode.
We aim to develop a method that can produce a program representing a policy that can be executed to maximize the discounted return,
\ie
\begin{math}
    \max_{\rho} \mathbb{E}_{a \sim \text{EXEC}(\rho)}[\sum_{t=0}^T \gamma^t
    r_t]
\end{math},
where EXEC$(\cdot)$ returns the actions induced by executing the program policy $\rho$ in the environment. 
This objective is a special case of the standard RL objective 
where a policy is represented as a program in a DSL 
and the policy rollout is obtained by executing the program. 

%% file: figure/dsl.tex
\begin{figure} 

\centering
\begin{mdframed}[font=\scriptsize]
\vspace{-0.4cm}
{
    \begin{align*}
    \text{Program}\ \rho &\coloneqq \text{DEF}\  \text{run}\ \text{m}(\ s\ \text{m})\\
    \text{Repetition} \ n &\coloneqq \text{Number of repetitions}\\
    \text{Perception} \ h & \coloneqq \text{frontIsClear} \ | \ \text{leftIsClear} \ | \  \text{rightIsClear} \ | \ \\ 
    & \text{markerPresent} \ | \ \text{noMarkerPresent} \\
    \text{Condition} \ b &\coloneqq \text{perception h} \ | \ \text{not} \ \text{perception h} \\
    \text{Action} \ a &\coloneqq \text{move} \ | \ \text{turnLeft}
    \ | \ \text{turnRight} \ | \ \\
    & \text{putMarker} \ | \ \text{pickMarker} \\
    \text{Statement}\ s &\coloneqq \text{while}\ \text{c}(\ b\ \text{c})\ \text{w}(\ s\ \text{w}) \ | \ s_1 ; s_2 \ | \ a \ | \\ 
    & \ \text{repeat}\ \text{R=}n\ \text{r}(\ s\ \text{r}) \ | \ \text{if}\ \text{c}(\ b\ \text{c})\ \text{i}(\ s\ \text{i}) \ | \\ 
    & \ \text{ifelse}\ \text{c}(\ b\ \text{c})\ \text{i}(\ s_1\ \text{i}) \  \text{else}\ \text{e}(\ s_2\ \text{e}) \\
    \end{align*}
\vspace{-0.9cm}
}
\end{mdframed}
    \vspace{-0.4cm}
    \caption[]{
        \small
        The domain-specific language (DSL) for the Karel domain, features an agent that can navigate through a grid world and interact with objects.
        \label{fig:dsl}
    }
\end{figure}

%% file: text/4_approach.tex

\vspacesection{Approach}
\label{sec:approach}

\input{figure/model}

Our goal is to design a framework that can synthesize task-solving programs based on the rewards obtained from MDPs.
We adapt the idea of learning a program embedding space to continuously parameterized a diverse set of programs proposed in LEAPS~\cite{trivedi2021learning}.
Then, instead of searching for a task-solving program in the learned program embedding space, our key insight is to learn a meta-policy that can hierarchically compose programs to form a more expressive task-solving program.
Our proposed framework, dubbed Hierarchical Programmatic Reinforcement Learning (HPRL), 
is capable of producing out-of-distributionally long and complex programs. 
Moreover, HPRL can make delicate adjustments to synthesized programs according to rewards obtained from the environment. 

\mysecref{sec:method_stage1} presents how LEAPS learns a program embedding space 
to continuously parameterize a set of randomly generated programs and describes our proposed procedure to produce a dataset containing more diverse programs.
Then, to reduce the dimension of the learned program embedding 
for more efficient meta-policy learning,
\mysecref{sec:method_compress} introduces how we compress the embedding space.
Finally, in \mysecref{sec:method_stage2}, we describe our method for learning a meta-policy, 
whose action space is the learned program embedding space, 
to hierarchically compose programs and yield a task-solving program.
An overview of our proposed framework is illustrated in~\myfig{fig:model}.

\vspacesubsection{Learning a Program Embedding Space}
\label{sec:method_stage1}

We aim to learn a program embedding space that continuously parameterizes a diverse set of programs. Moreover, a desired program embedding space should be behaviorally smooth, \ie programs that induce similar execution traces should be embedded closely to each other and programs with diverging behaviors should be far from each other in the embedding space.

To this end, we adapt the technique proposed in LEAPS~\cite{trivedi2021learning},
which trains
an encoder-decoder neural network architecture
on a pre-generated program dataset.
Specifically, a recurrent neural network program encoder $q_\phi$ learns to
encode a program $\rho$ (\ie sequences of program tokens) into a program embedding space, yielding a program embedding $v$; 
a recurrent neural network program decoder $p_\theta$ learns to decode
a program embedding $v$ to produce reconstructed programs $\hat{\rho}$.
The program encoder and the program decoder are trained to optimize 
the $\beta$-VAE~\cite{higgins2016beta} objective:
\begin{equation}
\label{eq:ploss} 
\begin{split}
\mathcal{L}_{\theta,\phi}^{\text{P}}(\rho) = -
\mathbb{E}_{v \sim q_\phi(v\vert\rho)} [\log
p_\theta(\rho\vert v)] \\ + \beta
D_\text{KL}(q_\phi(v\vert\rho)\|p_\theta(v)),
\end{split}
\end{equation}
where $\beta$ balances the reconstruction loss and the representation capacity of the program embedding space (\ie the latent bottleneck). 

To encourage behavioral smoothness, \citet{trivedi2021learning}
propose two additional objectives.
The \textit{program behavior reconstruction loss} minimizes 
the difference between 
the execution traces of the given program EXEC($\rho$)
and the execution traces of the reconstructed program EXEC($\hat{\rho}$).
On the other hand,
the \textit{latent behavior reconstruction loss} brings closer 
the execution traces of the given program EXEC($\rho$)
and the execution traces produced by feeding
the program embedding $v$ 
to a learned neural program executor $\pi(a|v, s)$:
\begin{equation}
\begin{split}
    \label{eq:lloss}
    \mathcal{L}_{\pi, \phi}^{\text{L}}(\rho, \pi) = - \mathbb{E}_{v \sim q_\phi(v\vert\rho)}[\sum_{t=1}^{H} \sum_{i=1}^{|\mathcal{A}|} \text{EXEC}_{t, i}(\rho) \\ \log \pi(a_i|v, s_t)],
\end{split}
\end{equation}
where $H$ denotes the horizon of EXEC($\rho$), 
$|\mathcal{A}|$ denotes the cardinality of the action space,
and $\text{EXEC}_{t, i}(\rho)$ is the boolean function indicating if the action equals to $a_i$ at time step $t$ while executing program $\rho$.

We empirically found that optimizing 
the \textit{program behavior reconstruction loss}
does not yield a significant performance gain.
Yet, due to the non-differentiability nature of program execution, optimizing this loss via REINFORCE~\citep{williams1992simple} is unstable. 
Moreover, performing on-the-fly program execution during training significantly slows down the learning process. 
Therefore, we exclude the \textit{program behavior reconstruction loss}, yielding 
our final objective for learning a program embedding space
as a combination of the $\beta$-VAE objective 
$\mathcal{L}_{\theta, \phi}^{\text{P}}$ and 
the \textit{latent behavior reconstruction loss}
$\mathcal{L}_\pi^{\text{L}}$:
\begin{equation}
    \label{eq:fullloss}
    \min_{\theta, \phi, \pi} 
    \mathcal{L}_{\theta, \phi}^{\text{P}}(\rho) + 
    \lambda \mathcal{L}_\pi^{\text{L}}(\rho, \pi),
\end{equation}
where $\lambda$ determines the relative importance of these losses.

\vspacesubsection{Compressing the Learned Program Embedding Space}
\label{sec:method_compress}

The previous section describes a method for constructing
a program embedding space that continuously parameterizes programs.
Next, given a task defined by an MDP, we aim to learn a meta-policy that predicts 
a sequence of program embeddings as actions 
to compose a task-solving program.
Hence, a low-dimensional program embedding space 
(\ie a smaller action space) is ideal for efficiently learning such a meta-policy.
Yet, to embed a large number of programs with diverse behaviors, 
a learned program embedding space needs to be extremely high-dimensional.

Therefore, our goal is to bridge the gap between a
high-dimensional program embedding space with sufficient 
representation capacity 
and a desired low-dimensional action space for learning 
a meta-policy.
To this end, we augment the encoder-decoder architecture with additional fully-connected layers to further reduce the program embedding space.
Specifically, we employ a compression encoder $f_\omega$ that takes the output of the program encoder $q_\phi$ as input and compresses it into a lower-dimensional program embedding $z$; 
also, we employ a compression decoder $g_\psi$
that takes a program embedding as input and decompresses it to produce a reconstructed higher-dimensional program embedding $\hat{v}$, which is then fed to the program decoder $p_\theta$ to produce a reconstructed program $\hat{\rho}$.

With this modification, the $\beta$-VAE objective and 
the \textit{latent behavior reconstruction loss} 
can be rewritten as:
\begin{equation}
\label{eq:compress_ploss} 
\begin{split}
\mathcal{L}_{\theta,\phi, \omega, \psi}^{\text{P}}(\rho) = -
\mathbb{E}_{z \sim f_\omega(z|q_\phi(\rho))} [\log
p_\theta(\rho|(g_\psi(z)))] \\ + \beta
D_\text{KL}(f_\omega(q_\phi(z\vert\rho))\|p_\theta(g_\psi(z))),
\end{split}
\end{equation}
and
\begin{equation}
\label{eq:compress_lloss}
\begin{split}
    \mathcal{L}_{\pi, \omega, \phi}^{\text{L}}(\rho, \pi) = - \mathbb{E}_{z \sim f_\omega(z|q_\phi(\rho))}[\sum_{t=1}^{H} \sum_{i=1}^{|\mathcal{A}|} \text{EXEC}_{t,i}(\rho) \\ \log \pi(a_i|z, s_t)],
\end{split}
\end{equation}
We train the program encoder $q_\phi$, 
the compression encoder $f_\omega$, 
the compression decoder $g_\psi$, 
the program decoder $p_\theta$, 
and the neural execution policy $\pi$ 
in an end-to-end manner as described in \myalgo{alg:algo_learning_emb}.
We discuss how the dimension of the program embedding space affects 
the quality of latent program embedding space in~\mysecref{sec:dim}.

\input{text/algo_learning_emb}

\vspacesubsection{Learning a Meta-Policy to Compose the Task-Solving Program}
\label{sec:method_stage2}

Once an expressive, smooth, yet compact program embedding space is learned, given a task described by an MDP, we propose to train a meta-policy $\pi_{\text{meta}}$ following \myalgo{alg:algo_learning_meta} to compose a task-solving program. 
Specifically, the learned program embedding space is used as a continuous action space for the meta-policy $\pi_{\text{meta}}$. 
We formulate the task of composing programs as a finite-horizon MDP whose horizon is $|H|$.
At each time step $i$, the meta-policy $\pi_{\text{meta}}$ takes an input state $s^i$ and predicts one latent program embedding $z_i$ as action, 
which can be decoded to its corresponding program $\rho_i$ using the learned compression decoder and program decoder $p_\theta(g_\psi(z_i))$. 
Then, the program $\rho_i$ is executed with $\text{EXEC}(\cdot)$ to interact with the environment for a period from $1$ to $T^i$,
yielding the cumulative reward $r^{i+1} = \sum_{t=1}^{T^i} r^i_t$ and the next state $s^{i+1} = [s^i_1, s^i_2, ..., s^i_{T_i}][-1]$ after the program execution. 
The operator $\cdot[-1]$ returns the last object in the sequence, and we take the last state of the program execution as the next macro input state, i.e.,  $s^{i+1} = [s^i_1, s^i_2, ..., s^i_{T_i}][-1] = s^i_{T_i}$.
A program will terminate when it is fully executed or once 100 actions have been triggered.
Note that the time steps $i$ considered here are macro time steps, each involves a series of state transitions and returns a sequence of rewards.
The environment will return the next state $s^{i+1}$ and cumulative reward $r^{i+1}$ to the agent to predict the next latent program embedding $z_{i+1}$.
The program composing process terminates after repeating $|H|$ steps.

The synthesized task-solving program $\mathcal{P}$ is obtained by 
sequentially composing the generated program 
$\langle\rho_i|i=1...|H|\rangle$, 
where $\langle \cdot \rangle$ denotes an operator 
that concatenates programs in order 
to yield a composed program.
Hence, the learning objective of the meta-policy $\pi_{\text{meta}}$ is to maximize the total cumulative return $\mathcal{J}_{\pi_{\text{meta}}}$:

\begin{equation}
    \label{eq:pi_meta_reward}
    \mathcal{J}_{\pi_{\text{meta}}} = \mathbb{E}_{\mathcal{P} \sim \pi_\textsc{meta}}[\sum_{i=1}^{|H|} \gamma ^ {i-1} \mathbb{E}_{a \sim \text{EXEC} (\rho_i)}[r^{i+1}]
    .
\end{equation}

where $\gamma$ is the discount factor for macro time steps MDP and $a$ is the primitive action triggered by $\text{EXEC}(\rho_i)$.
We detail the meta-policy training procedure in~\myalgo{alg:algo_learning_meta}.

\input{text/algo_learning_meta}

While this work formulates the program synthesis task as a finite-horizon MDP where a fixed number of programs $|H|$ are composed, 
we can instead learn a termination function that decides when to finish the program composition process, which is left to future work.

%% file: figure/model.tex
\begin{figure*}[ht]
\centering
    \includegraphics[width=\textwidth]{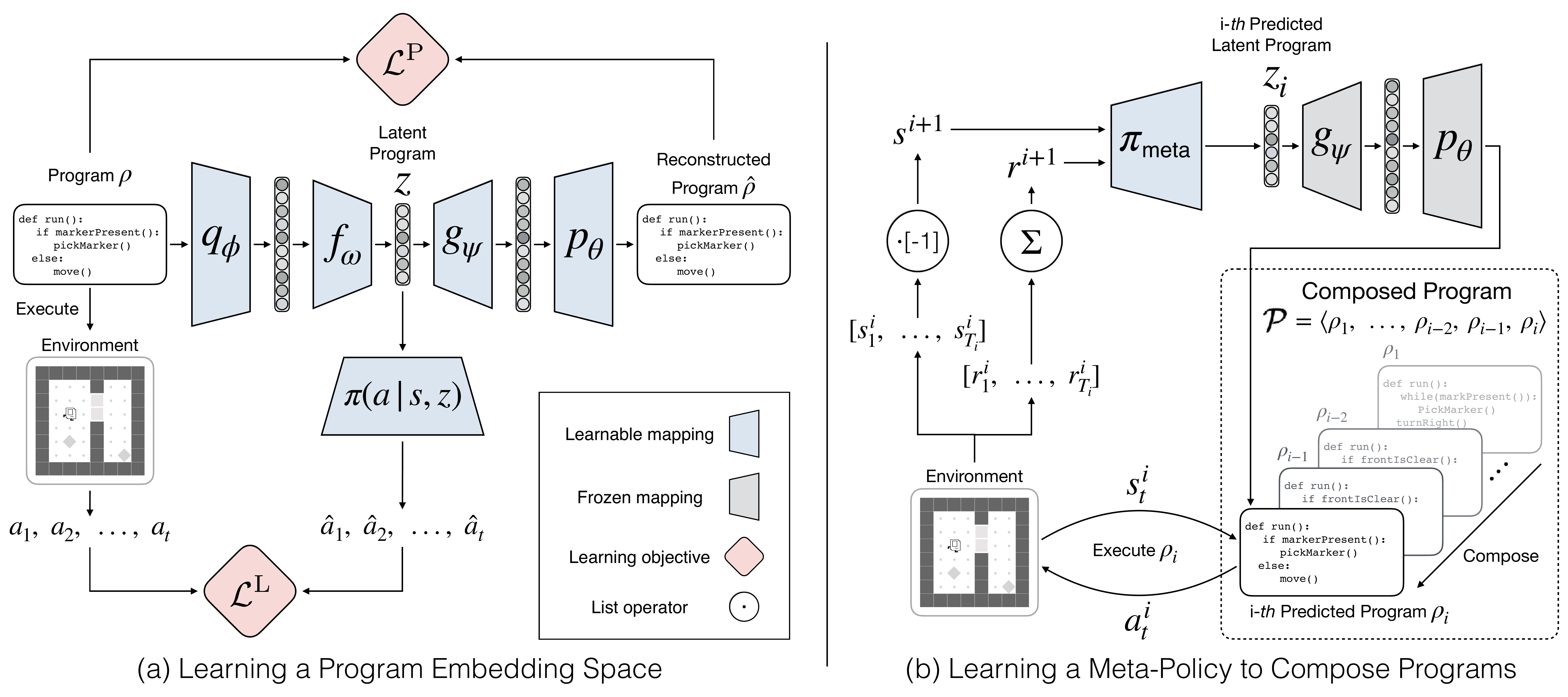}

    \caption[]{
        \small 
        \textbf{\myshorttitle{}.}
        \textbf{(a) Learning a Program Embedding Space}: a continuously parameterized latent program space can be learned using the program encoder $q_\phi$, decoder $p_\theta$, and a neural executor policy $\pi$
        by optimizing the two reconstruction objectives:
        $\mathcal{L}^{P}$ and $\mathcal{L}^{L}$.
        To reduce the dimensionality of the program embedding space to facilitate task learning,
        we employ a compression encoder $f_\omega$ and
        a compression decoder $g_\psi$.
        \textbf{(b) Learning a Meta-Policy to Compose Programs}: 
        given a task described by an MDP,
        we propose to train a meta-policy $\pi_{meta}$ 
        to compose a sequence of programs, and yield a task-solving program.
        Specifically, 
        at each macro time step $i$, 
        the meta-policy $\pi_{meta}$ 
        predicts a latent program embedding $z_i$,
        which can be decoded to the corresponding program $\rho_i = p_\theta(g_\psi(z_i))$. 
        We then execute the program $\rho_i$ in the environment, 
        which returns the cumulative reward $r^{i+1}$ 
        and the next state $s^{i+1}$ to the meta policy. 
        Then, the meta-policy can synthesize the next program $\rho_{i+1}$ based on $ s^{i+1}$.
        Upon termination, the predicted task-solving program is composed of a series of synthesized programs
        $\mathcal{P} = \langle\rho_1, \rho_2, ..., \rho_{|H|-1}, \rho_{|H|} \rangle$.
        \label{fig:model}
    }
\end{figure*}

%% file: text/algo_learning_emb.tex
\begin{algorithm}[H]
\caption{HPRL: Learning Latent Program Embedding Space}
\label{alg:algo_learning_emb}
\textbf{Input}: Program Dataset $D_{program}$, Training Epoch $N_{epoch}$\\
\textbf{Output}: Compression Decoder $g_\psi$, Program Decoder $p_\theta$
\begin{algorithmic}[1]
  \STATE Initialize the program encoder $q_\phi$ and decoder $p_\theta$, compression encoder $f_\omega$ and decoder $g_\psi$, neural execution policy $\pi$.
  \FOR{$\text{epoch}$ in range(1, $N_{epoch}$)}
    \FOR{program $\rho$ in $D_{program}$}
    \STATE    $z = f_\omega(q_\phi(\rho))$
    \STATE    $\hat{\rho} = p_\theta(g_\psi(z))$ 
    \STATE    Compute $\mathcal{L}_{total} = \mathcal{L}^P_{\theta, \phi, \omega, \phi}(\rho) + \lambda\mathcal{L}^L_{\pi, \omega, \phi}(\rho, \pi)$
    \STATE    Fit $\phi, \omega, \psi, \theta, \pi$ to minimize $\mathcal{L}_{total}$
    \ENDFOR
  \ENDFOR
\end{algorithmic}
\end{algorithm}

%% file: text/algo_learning_meta.tex
\begin{algorithm}[H]
\caption{HPRL: Meta-Policy Training} 
\label{alg:algo_learning_meta}
\textbf{Input}: Program Decoder $p_\theta$, Compression Decoder $g_\psi$, Meta-Policy Training Step $T_{meta}$, Horizon $|H|$\\
\textbf{Output}: Task Solving Program $\mathcal{P}$
\begin{algorithmic}[1]
  \STATE Initialize a meta-policy $\pi_{meta}$
  \STATE Load and fix $p_\theta, g_\psi$ for Meta-Policy Training
  \FOR{Training Episode in range(1, $T_{meta}/|H|$)}
    \STATE Receive initial state $s^1$ from the Karel environment
    \FOR{i in range(1, $|H|$)}
      \STATE $z_i = \pi_{meta}(s^i)$
      \STATE $\rho_i = p_\theta(g_\psi(z_i))$
      \STATE Interact with the environment by \textsc{EXEC}($\rho_i$)
      \STATE Receive $[s^i_1, ..., s^i_T]$ and $[r^i_1, ..., r^i_{T_i}]$
      \STATE $r^{i+1} = \sum_{t=1}^{T^i} r^i_t$
      \STATE $s^{i+1} = [s^i_1, s^i_2, ..., s^i_{T_i}][-1]$ 
    \ENDFOR
    \STATE Calculate $\mathcal{J}_{\pi_{meta}}$ based on the collected $\{(s^i, z^i, r^{i+1}, s^{i+1})|i=1, ..., |H|\}$
    \STATE Optimize $\pi_{meta}$ to maximize $\mathcal{J}_{\pi_{meta}}$
  \ENDFOR
\end{algorithmic}
\end{algorithm}

%% file: text/5_experiment.tex

\vspacesection{Experiments}
\label{sec:experiment}

We design and conduct experiments to compare our proposed framework (HPRL) to its variants and baselines.

\vspacesubsection{Karel domain}

\label{sec:karel}

\input{figure/karel_env_hard}

For the experiments and ablation studies, 
we adopt the Karel domain~\citep{pattis1981karel},
which is widely used in program synthesis 
and programmatic reinforcement learning~\citep{bunel2018leveraging,
shin2018improving, sun2018neural, chen2019executionguided, trivedi2021learning}.
The Karel agent in a gridworld can navigate and interact with objects (\ie markers). 
The action and perception are detailed in~\myfig{fig:dsl}. 

To evaluate the proposed framework and the baselines,
we consider two problem sets.
First, 
we use the \textsc{Karel} problem set proposed in~\cite{trivedi2021learning}, 
which consists of six tasks.
Then, we propose a more challenging set of tasks,
\textsc{Karel-Hard} problem set (shown in~\myfig{fig:karel_hard}),
which consists of four tasks.
In most tasks, initial configurations such as agent and goal locations, 
wall and marker placement, are randomly sampled upon every episode reset.

\myparagraph{{\textsc{Karel} Problem Set}}
The \textsc{Karel} problem set introduced in~\cite{trivedi2021learning} consists of six tasks: \textsc{StairClimber}, \textsc{FourCorner}, \textsc{TopOff}, \textsc{Maze}, \textsc{CleanHouse} and \textsc{Harvester}. 
Solving these tasks requires the following ability.
\textit{Repetitive Behaviors}: to conduct the same behavior for several times, \ie placing markers on all corners (\textsc{FourCorner}) or move along the wall (\textsc{StairClimber}). 
\textit{Exploration}: to navigate the agent through complex patterns (\textsc{Maze}) or multiple chambers (\textsc{CleanHouse}).
\textit{Complexity}: to perform specific actions, \ie put markers on marked grid (\textsc{TopOff}) or pick markers on marked grid (\textsc{Harvester}).
For further description about the \textsc{Karel} problem set, please refer to \mysecref{sec:app_karel}.

\myparagraph{\textsc{Karel-Hard} Problem Set} We design a more challenging set of tasks, the \textsc{Karel-Hard} problem set. The ability required to solve the tasks in this problem set can be categorized as follows: \textit{Two-stage exploration}: to explore the environment under different conditions, \ie pick up the marker in one chamber to unlock the door, and put the marker in the next chamber (\textsc{DoorKey}). \textit{Additional Constraints}: to perform specific actions under restrictions, \ie traverse the environment without revisiting the same position (\textsc{OneStroke}), place exactly one marker on all grids (\textsc{Seeder}), and traverse the environment without hitting a growing obstacle (\textsc{Snake}). More details about the \textsc{Karel-Hard} problem set can be found in \mysecref{sec:app_karel_hard}.

\vspacesubsection{Experimental Settings}
\label{sec: settings}
\mysecref{sec:program_dataset_generation} introduces 
the procedure for generating the program dataset used for learning a program embedding space. 
The implementation of the proposed framework is described in~\mysecref{sec:HPRL_implementation}.

\vspacesubsubsection{Karel DSL Program Dataset Generation with Our Improved Generation Procedure}
\label{sec:program_dataset_generation}
The Karel program dataset used in this work includes one million programs. All the programs are generated based on syntax rules of the Karel DSL with a maximum length of $40$ program tokens.
While \citet{trivedi2021learning} randomly sample to generate program sequences, 
we propose an improved program generation procedure as follows.
We filter out counteracting programs (\eg termination state equals initial state after program execution), repetitive programs (\eg programs with long common sub-sequences) and programs with canceling action sequences (e.g., \texttt{turnLeft} followed by \texttt{turnRight}).
These rules significantly improve the diversity and expressiveness of the generated programs and induce a more diverse and complex latent program space.
More details can be found in~\mysecref{sec:app_datasets}.

\vspacesubsubsection{Implementation}
\label{sec:HPRL_implementation}

\myparagraph{Encoders \& Decoders}
We use GRU~\citep{Cho2014GRU} layer to implement both the program encoder $q_\phi$ and the program decoder $p_\theta$ mentioned in~\mysecref{sec:method_stage1} with a hidden dimension of $256$. The last hidden state of the encoder $q_\phi$ is taken as the uncompressed program embedding $v$. This program embedding $v$ can be further compressed to a $64$-dimensional program embedding $z$ using the compression encoder $f_\omega$ and compression decoder $g_\psi$ constructed by the fully-connected neural network as described in~\mysecref{sec:method_compress}.

\myparagraph{Neural Program Executor}
The neural program executor $\pi$ is implemented as a recurrent conditional policy $\pi(\cdot|z, s)$ using GRU layers, which takes the abstract state $s$ and the program embedding $z$ at each time step as input and predicts the execution trace.

\myparagraph{Meta-Policy}
To implement the meta-policy $\pi_{\text{meta}}$, we use convolutional layers~\cite{fukushima1982neocognitron, krizhevsky2017imagenet} to extract features from the Karel states and then process them with GRU layers for predicting program embeddings. To optimize the meta-policy, we use two popular reinforcement learning algorithms, PPO \citep{schulman2017proximal} and SAC \citep{haarnoja18b}, and report their experimental results as HPRL-PPO and HPRL-SAC, respectively.

More details on hyperparameters, training procedure, and implementation can be found in~\mysecref{sec:app_hyperparameters}.

\input{table/karel}

\input{table/karel_hard}

\vspacesubsubsection{Baseline Approaches}

We compare HPRL with the following baselines.

\begin{itemize}
    \item \myparagraph{Best-sampled} 
    It randomly samples $1000$ programs from the learned program embedding space and reports the highest return achieved by the sampled programs. 

    \item \myparagraph{DRL and DRL-abs} Deep RL baselines from~\cite{trivedi2021learning}.
    DRL observes a raw state (grids) input from the Karel environment,
    while DRL-abs is a recurrent neural network policy that takes abstracted state vectors from the environment as input. The abstracted state vectors consist of binary values of the current state (\eg [\texttt{frontIsClear() == True}, \texttt{markerPresent()==False, ...}]).

    \item \myparagraph{VIPER} A programmatic RL method proposed by \citet{bastani2018verifiable}. It uses a decision tree to imitate the behavior of a learned DRL policy.
    
    \item  \myparagraph{LEAPS} A programmatic RL framework proposed by \citet{trivedi2021learning} that uses Cross-Entropy Method \citep{rubinstein1997optimization} to search task-solving program in a learned continuous program embedding space.

    \item \myparagraph{LEAPS-ours} The LEAPS framework trained on the proposed Karel program dataset described in~\mysecref{sec:program_dataset_generation}. This is used to compare our proposed program dataset generation procedure with the generation approach used in~\cite{trivedi2021learning}.
\end{itemize}
   
More details of these baselines can be found in~ \mysecref{sec:app_baseline}.

\vspacesubsection{Experimental Results}
\label{sec:results}
We evaluate the cumulative return of all methods on the \textsc{Karel} problem set and the \textsc{Karel-Hard} problem set.
The results of the two problem sets are presented in \mytable{table:karel} and \mytable{table:karel_hard}, respectively. 
The range of the cumulative return is within $[0, 1]$ on all tasks.
\mysecref{sec:app_problem_set} describes the detailed definition of the reward function for each task. 
The performance of DRL, DRL-abs, VIPER, and LEAPS reproduced with the implementation provided by \citet{trivedi2021learning}. 
The average cumulative return and standard deviation of all the methods on each task are evaluated over five random seeds to ensure statistical significance. 

\myparagraph{Overall \textsc{Karel} Performance} 
The experimental results on \mytable{table:karel} show that HPRL-PPO outperforms all other approaches on all tasks. Furthermore, HPRL-PPO can completely solve all the tasks in the \textsc{Karel} problem set. 
The Best-sampled results justify the quality of the learned latent space as tasks like \textsc{StairClimber} and \textsc{maze} can be entirely solved by one (or some) of 1000 randomly sampled programs. However, all 1000 randomly sampled programs fail on tasks that require long-term planning and exploration (e.g., \textsc{FourCorner, CleanHouse} and \textsc{Harvester}), showing the limit of the simple search-based method. 
On the other hand, we observe that LEAPS-ours outperforms LEAPS on all of the six tasks in the \textsc{Karel} problem set, showing that the proposed program generation process helps improve the quality of the program embedding space and leads to the better program search result.

\myparagraph{Overall \textsc{Karel-Hard} Performance} 
To further testify the efficacy of the proposed method, we evaluate Best-sampled, LEAPS, LEAPS-ours, HPRL-PPO, and HPRL-SAC on the \textsc{Karel-Hard} problem set. HPRL-PPO outperforms other methods on \textsc{OneStroke} and \textsc{Seeder}, while all approaches perform similarly on \textsc{DoorKey}. The complexity of \textsc{OneStroke}, \textsc{Seeder}, and \textsc{Snake} makes it difficult for Best-sampled and LEAPS to find sufficiently long and complex programmatic policies that may not even exist in the learned program embedding space. In contrast, HPRL-PPO addresses this by composing a series of programs to increase the expressiveness and perplexity of the synthesized program. We also observe that LEAPS-ours achieve better performance than LEAPS, further justifying the efficacy of the proposed program generation procedure.

\myparagraph{PPO vs. SAC}
HPRL-SAC can still deliver competitive performance in comparison with HPRL-PPO. However, we find that HPRL-SAC is more unstable on complex tasks (\eg \textsc{TopOff}) and tasks with additional constraints (\eg \textsc{Seeder}). 
On the other hand, HPRL-PPO is more stable across all tasks and achieves better performance on both problem sets. 
Hence, we adopt HPRL-PPO as our main method in the following experiments.

\input{figure/karel_program_example}

\myparagraph{Qualitative Results}
\myfig{fig:karel_hard_program_examples_snake_table} presents the programs synthesized by LEAPS, LEAPS-ours, and HPRL-PPO. 
The programs produced by LEAPS 
are limited by the maximum length and the complexity of the pre-collected program dataset. 
On the other hand, HPRL-PPO can compose longer and more complex programs by synthesizing and concatenating shorter and task-oriented programs 
due to its hierarchical design.
More examples of synthesized programs can be found in~\mysecref{sec:app_generated_program_samples}.

\vspacesubsection{Additional Experiments}
\label{sec:additional_exp}

This section investigates (1) whether LEAPS~\citep{trivedi2021learning} and our proposed framework can synthesize out-of-distributional programs, (2) the necessity of the proposed compression encoder and decoder, and (3) the effectiveness of learning from dense rewards made possible by the hierarchical design of our framework.

\vspacesubsubsection{Synthesizing Out-of-Distributional Programs}
\label{sec:recon}

Programs that LEAPS can produce are fundamentally limited by the distribution of the program dataset since it searches for programs in the learned embedding space. More specifically, it is impossible for LEAPS to synthesize programs that are significantly longer than the programs provided in the dataset. This section aims to empirically verify this hypothesis and evaluate the capability of generating out-of-distributional programs. We create a set of target programs of lengths 25, 50, 75, and 100, each consisting of primitive actions (\eg \texttt{move}, \texttt{turnRight}). Then, we ask LEAPS and HPRL to fit each target program based on how well the program produced by the two methods can reconstruct the behaviors of the target program. The reconstruction performance is calculated as one minus the normalized Levenshtein Distance between the state sequences from the execution trace of the target program and from the execution trace of the synthesized program.
The result is presented in~\mytable{table:recon}.

HPRL consistently outperforms LEAPS with varying target program lengths, and the gap between the two methods grows more significant when the target program becomes longer. We also observe that the reconstruction score of LEAPS drops significantly as the length of target programs exceeds $40$, which is the maximum program length of the program datasets. This suggests that HPRL can synthesize out-of-distributional programs. 
Note that the performance of HPRL can be further improved when setting the horizon of the meta-policy $|H|$ to a larger number. Yet, for this experiment, we fix it to $5$ to better analyze our method.
More details (\eg evaluation metrics) can be found in~\mysecref{sec:app_recon_details}. 

\input{table/recon}

\input{table/ablation_dim}

\vspacesubsubsection{Dimensionality of Program Embedding Space}
\label{sec:dim}
Learning a higher-dimensional program embedding space
can lead to better optimization in the program reconstruction loss (\myeq{eq:compress_ploss}) and the latent behavior reconstruction loss (\myeq{eq:compress_lloss}).
Yet, learning a meta-policy in a higher-dimensional action space
can be unstable and inefficient.
To investigate this trade-off and 
verify our contribution of employing 
the compression encoder $f_\omega$ and compression decoder $g_\psi$, we experiment with
various dimensions of program embedding space and report the result in~\mytable{table:ablation_dim}.

The reconstruction accuracy measures whether learned encoders and decoders can perfectly reconstruct an input program or its execution trace.
The task performance evaluates the return achieved in \textsc{CleanHouse} and \textsc{Seeder} since they are considered more difficult from each problem set.
The result indicates that a $64$-dimensional program embedding space achieves satisfactory reconstruction accuracy and performs the best on the tasks.
Therefore, we adopt this (\ie $dim(z)=64$) for HPRL on all the tasks.

\input{figure/eps}

\vspacesubsubsection{Learning from Episodic Reward}
\label{sec:sparse}

We design our framework to synthesize a sequence of programs, allowing for accurately rewarding correct programs and penalizing wrong programs (\ie better credit assignment) with dense rewards. In this section, we design experiments to investigate the effectiveness of this design. To this end, instead of receiving a reward for executing each program (\ie \textit{dense}) in the environment, we modify \textsc{CleanHouse} and \textsc{Seeder} so that they only return cumulative rewards after all $|H|$ programs have been executed (\ie \textit{episodic}). The learning performance is shown in~\myfig{fig:eps_environment}, demonstrating that learning from \textit{dense} rewards yields better sample efficiency compared to learning from \textit{episodic} rewards. This performance gain is made possible by the hierarchical design of HPRL, which can better deal with credit assignment. In contrast, LEAPS~\cite{trivedi2021learning} is fundamentally limited to learning from episodic rewards.

%% file: figure/karel_env_hard.tex
\begin{figure}
    \centering
    \begin{subfigure}[b]{0.32\linewidth}
    \centering
    \includegraphics[trim=0 120 0 120, clip,height=\textwidth]{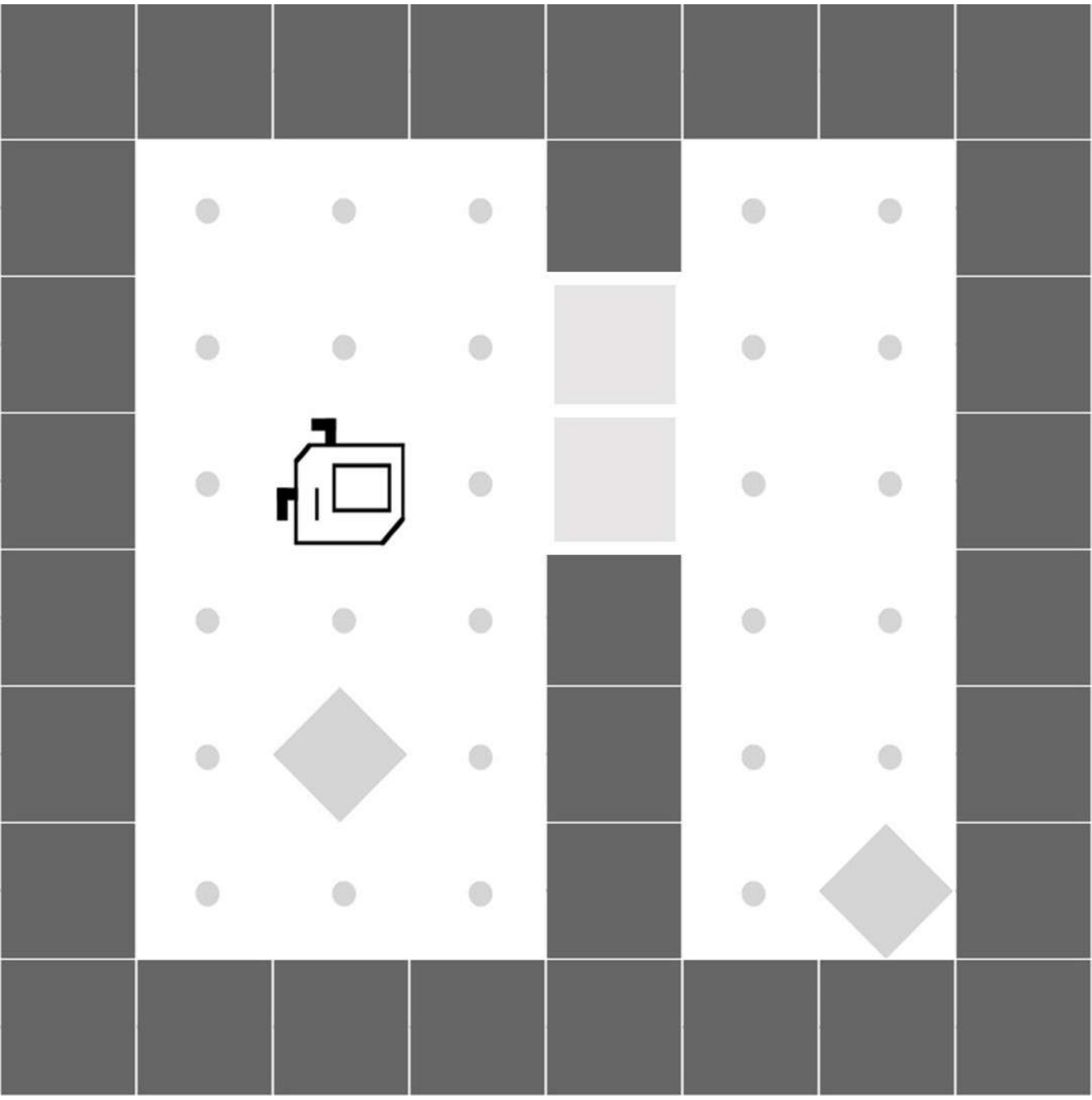}
    \SmallCaption{\textsc{DoorKey}}
    \end{subfigure}
    \begin{subfigure}[b]{0.32\linewidth}
    \centering
    \includegraphics[trim=0 120 0 120, clip,height=\textwidth]{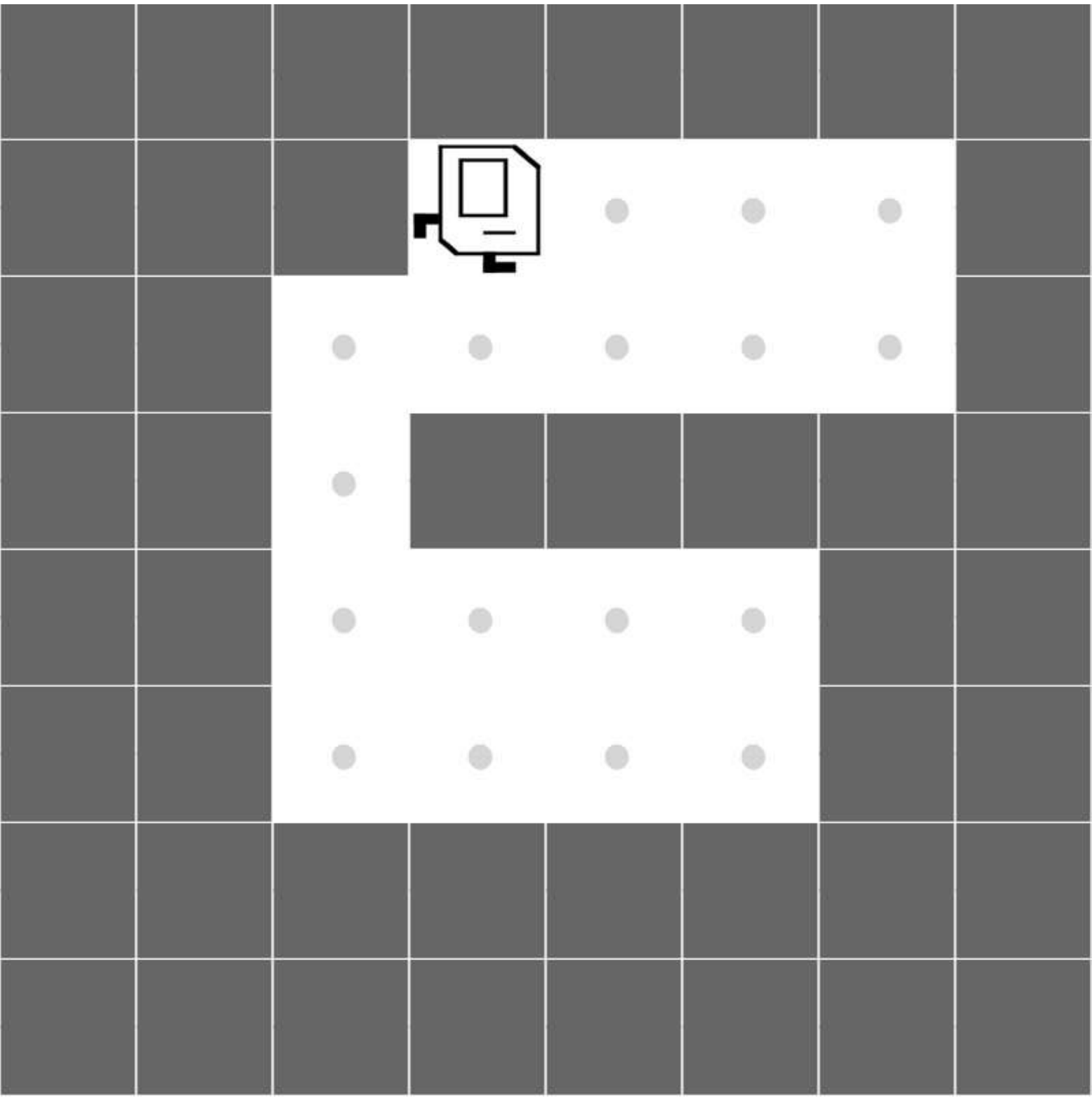}
    \SmallCaption{\textsc{OneStroke}}
    \end{subfigure}
    \\
    \begin{subfigure}[b]{0.32\linewidth}
    \centering
    \includegraphics[trim=0 120 0 120, clip,height=\textwidth]{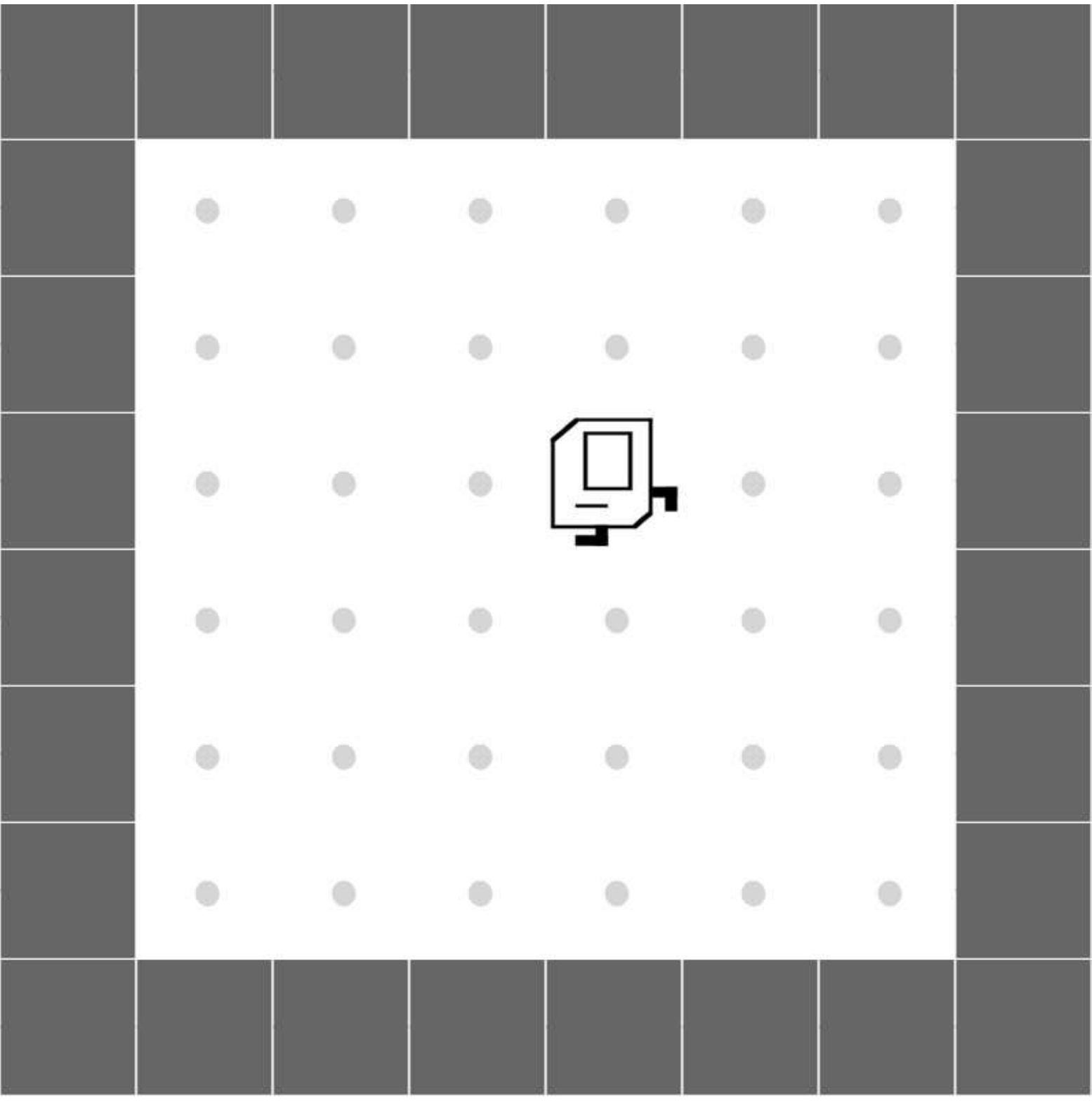}
    \SmallCaption{\textsc{Seeder}}
    \end{subfigure}
    \begin{subfigure}[b]{0.32\linewidth}
    \centering
    \includegraphics[trim=0 120 0 120, clip,height=\textwidth]{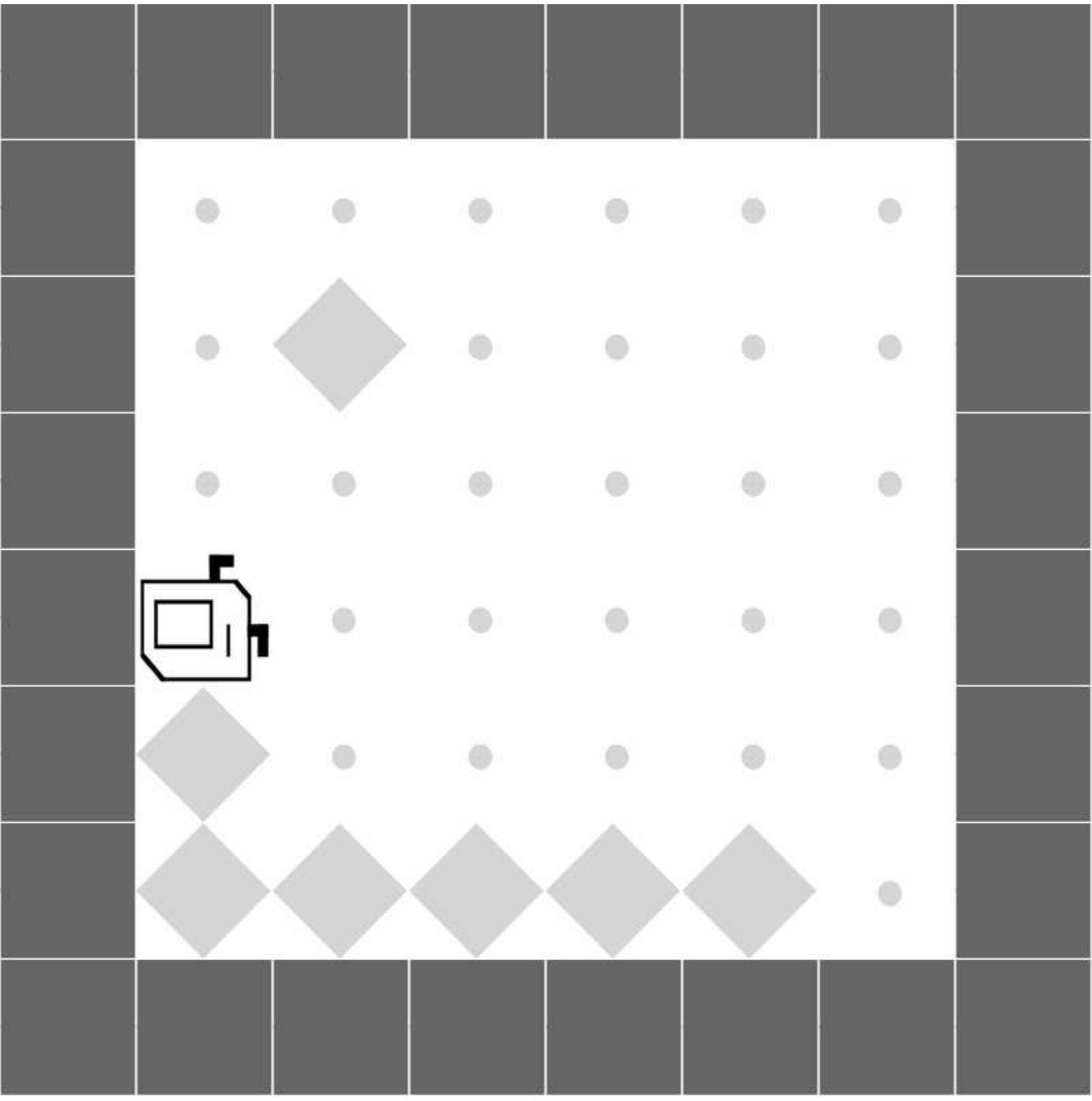}
    \SmallCaption{\textsc{Snake}}
    \end{subfigure}
    \caption[]{
    \textbf{\textsc{Karel-Hard} Problem Set}: 
    The four tasks require an agent to acquire
    a set of diverse, goal-oriented, and programmatic behaviors. 
    This is strictly more challenging compared to the \textsc{Karel} problem set proposed in~\cite{trivedi2021learning}.
    }
    \label{fig:karel_hard}
\end{figure}

%% file: table/karel.tex
\begin{table*}
\centering
\caption[]{Mean return and standard deviation of all methods across the \textsc{Karel} problem set, 
evaluated over five random seeds. HPRL-PPO outperforms all prior approaches and achieves the maximum score on all tasks. HPRL-SAC completely solves four out of six tasks.}
\scalebox{0.8}{\begin{tabular}{@{}ccccccc@{}}\toprule
Method & \textsc{StairClimber} & \textsc{FourCorner} & \textsc{TopOff} & \textsc{Maze} & \textsc{CleanHouse} & \textsc{Harvester}\\
\cmidrule{2-7}
Best-Sampled & \textbf{1.00} $\pm$ 0.00  & 0.25 $\pm$ 0.00  & 0.60 $\pm$ 0.07  & \textbf{1.00} $\pm$ 0.00    & 0.05 $\pm$ 0.04     & 0.17 $\pm$ 0.00    \\ 
DRL & \textbf{1.00} $\pm$ 0.00 & 0.29 $\pm$ 0.05 & 0.32 $\pm$ 0.07 & \textbf{1.00} $\pm$ 0.00 & 0.00 $\pm$ 0.00 & 0.90 $\pm$ 0.10\\
DRL-abs & 0.13 $\pm$ 0.29 & 0.36 $\pm$ 0.44 & 0.63 $\pm$ 0.23 & \textbf{1.00} $\pm$ 0.00 & 0.01 $\pm$ 0.02 & 0.32 $\pm$ 0.18 \\
VIPER & 0.02 $\pm$ 0.02 & 0.40 $\pm$ 0.42 & 0.30 $\pm$ 0.06 & 0.69 $\pm$ 0.05 & 0.00 $\pm$ 0.00 & 0.51 $\pm$ 0.07\\
LEAPS & \textbf{1.00} $\pm$ 0.00 & 0.45 $\pm$ 0.40 & 0.81 $\pm$ 0.07 & \textbf{1.00} $\pm$ 0.00 & 0.18 $\pm$ 0.14 & 0.45 $\pm$ 0.28 \\
LEAPS-ours & \textbf{1.00} $\pm$ 0.00 & 0.50 $\pm$ 0.47 & 0.82 $\pm$ 0.11 & \textbf{1.00} $\pm$ 0.00 & 0.28 $\pm$ 0.27 & 0.82 $\pm$ 0.16 \\
\midrule
\method\\-SAC & \textbf{1.00} $\pm$ 0.00 & \textbf{1.00} $\pm$ 0.00 & 0.61 $\pm$ 0.25 & \textbf{1.00} $\pm$ 0.00 & 0.99 $\pm$ 0.02 & \textbf{1.00} $\pm$ 0.00 \\
\method\\-PPO & \textbf{1.00} $\pm$ 0.00 & \textbf{1.00} $\pm$ 0.00 & \textbf{1.00} $\pm$ 0.00 & \textbf{1.00} $\pm$ 0.00 & \textbf{1.00} $\pm$ 0.00 & \textbf{1.00} $\pm$ 0.00 \\
\bottomrule
\end{tabular}}
\label{table:karel}
\end{table*}

%% file: table/karel_hard.tex
\begin{table}
\centering
\caption[]{Mean return and standard deviation of all methods across the \textsc{Karel-Hard} problem set, 
evaluated over five random seeds. HPRL-PPO achieves best performance across all tasks.}
\scalebox{0.80}{\begin{tabular}{@{}ccccc@{}}\toprule
Method 
& \textsc{DoorKey} & \textsc{OneStroke} & \textsc{Seeder} & \textsc{Snake}\\
\cmidrule{2-5}
Best-Sampled &  \textbf{0.50} $\pm$ 0.00 &  0.55 $\pm$ 0.34 &  0.17 $\pm$ 0.00 &  0.20 $\pm$ 0.00 \\ 
LEAPS 
& \textbf{0.50} $\pm$ 0.00 & 0.65 $\pm$ 0.19 & 0.51 $\pm$ 0.21 & 0.21 $\pm$ 0.15 \\
LEAPS-ours 
& \textbf{0.50} $\pm$ 0.00 & 0.72 $\pm$ 0.06 & 0.57 $\pm$ 0.02 & 0.25 $\pm$ 0.07 \\
\midrule
{\method\\-SAC}
& \textbf{0.50} $\pm$ 0.00 & 0.76 $\pm$ 0.05 & 0.27 $\pm$ 0.10 & \textbf{0.28} $\pm$ 0.15 \\
{\method\\-PPO} 
& \textbf{0.50} $\pm$ 0.00 & \textbf{0.80} $\pm$ 0.02 & \textbf{0.58} $\pm$ 0.07 & \textbf{0.28} $\pm$ 0.11 \\
\bottomrule
\end{tabular}
}
\label{table:karel_hard}
\end{table}

%% file: figure/karel_program_example.tex
\begin{figure}
\centering
\begin{mdframed}
\begin{subfigure}[t]{1.0\textwidth}
\centering
{
\begin{subfigure}[t]{0.49\textwidth}
LEAPS
\begin{lstlisting}
DEF run m( 
    turnRight 
    turnLeft 
    pickMarker 
    move 
    move 
    move 
    WHILE c( rightIsClear c) w( 
        turnLeft 
        move 
        move 
        w) 
    turnLeft 
    turnLeft 
    turnLeft 
    turnLeft 
    m) 
\end{lstlisting}

LEAPS-ours
\begin{lstlisting}
DEF run m( 
    move 
    turnRight 
    pickMarker 
    pickMarker 
    WHILE c( rightIsClear c) w( 
        turnLeft 
        move 
        move 
        w) 
    turnRight 
    move 
    move 
    move 
    m)
\end{lstlisting}

\end{subfigure}
\begin{subfigure}[t]{0.49\textwidth}
HPRL-PPO
\begin{lstlisting}
DEF run m( 
    move 
    WHILE c( noMarkersPresent c) w( 
        move 
        move 
        turnLeft 
        w) 
    move 
    turnLeft 
    m)
DEF run m( 
    move 
    WHILE c( noMarkersPresent c) w( 
        move 
        move 
        turnLeft 
        w) 
    m)
DEF run m( 
    move 
    WHILE c( noMarkersPresent c) w( 
        move 
        move 
        turnLeft 
        w) 
    move 
    turnLeft 
    m)
\end{lstlisting}
\end{subfigure}
}
\end{subfigure}
\end{mdframed}
\caption[Synthesized Programs on \textsc{Snake}]{\textbf{Synthesized Programs on \textsc{Snake}.}
HPRL synthesizes a longer and more complex program, outperforming LEAPS.
}
\label{fig:karel_hard_program_examples_snake_table}
\end{figure}

%% file: table/recon.tex
\begin{table}
\centering
\caption[]{\textbf{Learning to synthesize out-of-distributional programs.}
HPRL demonstrates superior performance in synthesizing 
out-of-distributionally long programs compared to LEAPS.
The gap between the two methods grows more significant when the length of the target program increases. 
}
\scalebox{0.8}{\begin{tabular}{@{}cccccc@{}}\toprule
\multirow{2}{*}{Method}& \multicolumn{4}{c}{Program Reconstruction Performance}\\
 & Len 25 & Len 50 & Len 75 & Len 100\\
\cmidrule{2-5}
LEAPS & 0.59 (0.14) &  0.31 (0.10) &  0.20 (0.05) &  0.13 (0.08)\\
\method{} & \textbf{0.60} (0.03) &  \textbf{0.36} (0.03) &  \textbf{0.29} (0.03) &  \textbf{0.26} (0.02)\\
\midrule
Improvement & 1.69\% & 16.13\% & 45.0\% & 100.0\% \\
\bottomrule
\end{tabular}}
\label{table:recon}
\end{table}

%% file: table/ablation_dim.tex
\begin{table}
\centering
\caption[]{\textbf{Dimensionality of the Program Embedding Space.} 
The 64-dimensional program embedding space demonstrates 
the best task performance with satisfactory reconstruction results.
}
\scalebox{0.8}{\begin{tabular}{@{}ccccc@{}}
\toprule
\multirow{2}{*}{dim($z$)} & \multicolumn{2}{c}{Reconstruction} & 
\multicolumn{2}{c}{Task Performance} \\
& Program & Execution & 
\textsc{CleanHouse} & \textsc{Seeder}\\
\cmidrule{2-5}
16   & 81.70\% & 63.21\% & 
0.47 (0.06) & 0.21 (0.02) \\
32   & 94.46\% & 86.00\% & 
0.84 (0.27) & 0.35 (0.16) \\
64   & 97.81\% & 95.58\% & 
\textbf{1.00} (0.00) & \textbf{0.58} (0.07) \\
128  & 99.12\% & 98.76\% & 
1.00 (0.00) & 0.57 (0.03) \\
256  & 99.65\% & 99.11\% & 
1.00 (0.00) & 0.55 (0.11) \\
\bottomrule
\end{tabular}}
\label{table:ablation_dim}
\end{table}

%% file: figure/eps.tex
\begin{figure}[t]
    \centering
    \begin{subfigure}[b]{0.49\linewidth}
    \label{fig:env_maze}
    \centering
    \includegraphics[width=\textwidth]{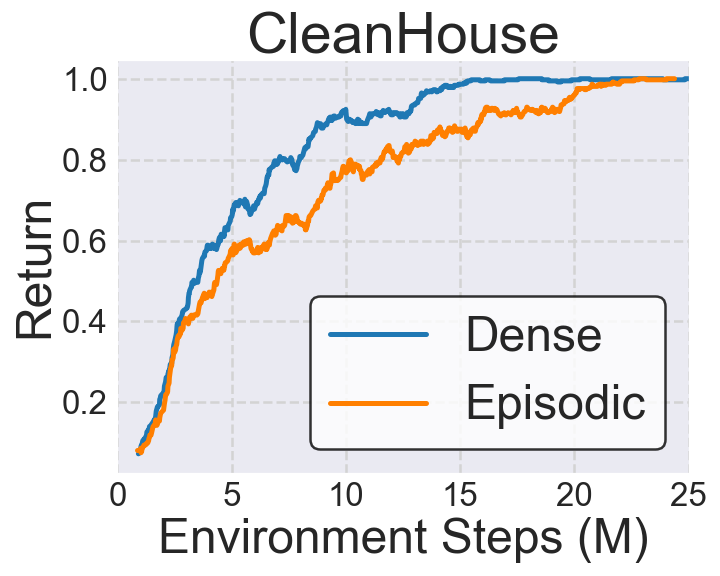}
    \end{subfigure}
    \begin{subfigure}[b]{0.49\linewidth}
    \label{fig:env_pick}
    \centering
    \includegraphics[width=\textwidth]{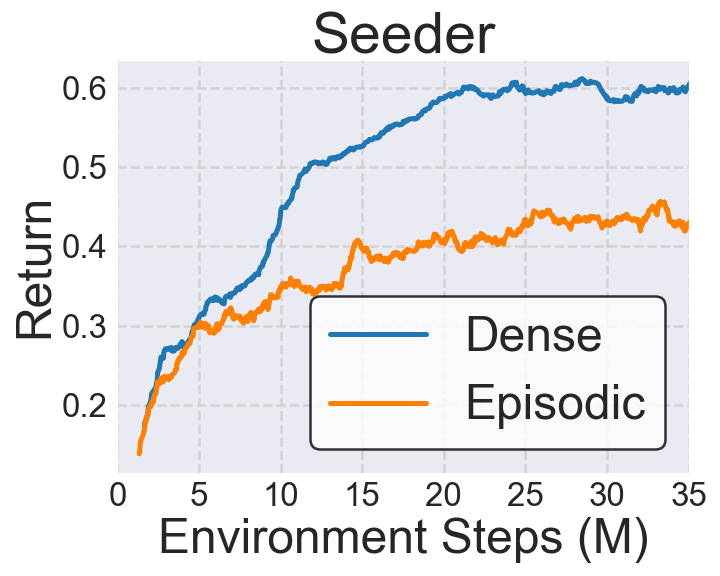}
    \end{subfigure}
    \caption{\textbf{Learning from Episodic Reward.} 
    We compare learning from dense and episodic rewards in \textsc{CleanHouse} and \textsc{Seeder}. Learning from dense rewards achieves better sample efficiency in both tasks, which is made possible by the hierarchical design of our proposed framework.
    }
    \label{fig:eps_environment}
\end{figure}    

%% file: text/6_conclusion.tex

\vspacesection{Conclusion}
\label{sec:conclusion}

We propose a hierarchical programmatic reinforcement learning framework (HPRL), which re-formulates solving a reinforcement learning task as synthesizing a task-solving program that can be executed to interact with the environment and maximize the return. Specifically, we first learn a program embedding space that continuously parameterizes a diverse set of programs generated based on our proposed program generation procedure. 
Then, we train a meta-policy, whose action space is the learned program embedding space, to produce a series of programs (\ie predict a series of actions) to yield a composed task-solving program. 
Experimental results in the Karel domain demonstrate that HPRL consistently outperforms baselines by large margins. 
Ablation studies justify our design choices, including the RL algorithms for learning the meta-policy and the dimensionality of the program embedding space. 
Additional experimental results confirm the two fundamental limitations of LEAPS~\cite{trivedi2021learning} and attribute the superior performance of HPRL to addressing these limitations.

%% file: text/7_ack.tex

\section*{Acknowledgement}
\label{sec:ack}

This work was supported by the National Taiwan University and its Department of Electrical Engineering, Graduate Institute of Networking and Multimedia, Graduate Institute of Communication Engineering, and College of Electrical Engineering and Computer Science. 
The authors also appreciate the fruitful discussion with the members of NTU Robot Learning Lab.

%% file: text/appendix.tex
\section*{Appendix}

\section{Discussion}

\subsection{Synthesizing Programs with Large-Language Models}
\label{sec:app_discussion_LLM}

Recently, leveraging large-language models (LLM) for program generation tasks has received increasing attention. For example, OpenAI Codex \citep{chen2021openaicodex}, AlphaCode \citep{li2022alphacode}, and related works~\citep{chen2023teaching, poesia2022synchromesh, sun2020treegen, jain2022jigsaw} aim to produce Python/C++ code from natural language descriptions. 
In contrast, our goal is to synthesize programs that describe the behaviors of learning agents from rewards. 

On the other hand, \citet{codeaspolicies2022} utilize LLMs to generate behavioral instructions given task-oriented prompts. 
However, their problem formulation involves human interventions to provide task-specific prompts, which significantly deviates from our problem formulation, which is to synthesize task-solving programs purely based on rewards automatically.  

Lastly, fine-tuning LLMs trained on code, such as CodeT5~\citep{wang2021codet5} and CodeT5+~\citep{wang2023codet5plus} models, on Karel programs and leveraging its embedding space as the action space is another promising way to further enhance the performance of the proposed method.

\subsection{Limitations}
\label{sec:app_scope_future_work}

\myparagraph{Karel Domain DSL} In this work, we choose the Karel DSL (\myfig{fig:dsl}) due to its popularity among prior works \citep{bunel2018leveraging, shin2018improving, sun2018neural, chen2019executionguided, trivedi2021learning} and the expressiveness of the behavioral description for the agent. Incorporating the proposed HPRL framework for synthesizing programs with a more general DSL is worth further exploring.

\myparagraph{Deterministic Environments} 
The Karel domain features a deterministic environment,
where any action has a single guaranteed effect without the possibility of failure or uncertainty.
Designing a DSL and a program synthesis framework
that can synthesize task-solving programs in 
stochastic environments is another promising research direction.

\section{Method Details}
\label{sec:app_baseline}

The details of each method are described in this section.

\subsection{\method{}}
The overall framework of HPRL consists of two parts: the pre-trained decoder, as mentioned in \mysecref{sec:method_stage1}, and the meta-policy described in \mysecref{sec:method_stage2}. The decoder is constructed with a one-layer unidirectional GRU, where the hidden size and input size are set to 256. Additionally, we employ a compression decoder to further compress the latent program space. This is achieved by using a fully-connected linear neural network with an output dimension of 256 and a latent embedding dimension of $[16, 32, 64, 128]$. Please note that the VAE with 256-dimensions program embedding space does not include the fully connected linear neural network.
The meta-policy neural network consists of a CNN neural network as a state feature extractor and a fully-connected linear layer for the action and value branch. The CNN neural network includes two convolutional layers. The filter size of the first convolutional layer is $32$ with $4$ channels, and the filter size of the second convolutional layer is $32$ with $2$ channels. The output of the state embedding is flattened into a vector of the same size as the output action vector. 


\subsection{DRL}
The DRL method implements a deep neural network trained using the Proximal Policy Optimization (PPO) algorithm for $2M$ time steps. It learns a policy that takes raw states (grids) from the Karel environment as input and predicts the next action. The raw state is represented by a binary tensor that reflects the state of each grid.

\subsection{DRL-abs}
DRL-abs utilizes a deep neural network with a recurrent policy and is trained using the PPO algorithm, which has demonstrated better performance compared to the Soft Actor-Critic (SAC) algorithm. It is also trained for $2M$ time steps. Instead of using the raw states (grids) of Karel, it takes abstract states as input. These abstract states are represented by a binary vector that encompasses all returned values of perceptions of the current state, \eg [\texttt{frontIsClear() == True}, \texttt{leftIsClear() == False}, \texttt{rightIsClear() == True}, \texttt{markerPresent() == False}, \texttt{noMarkersPresent() == True}].

\subsection{VIPER}
VIPER is a programmatic RL framework proposed by \citet{bastani2018verifiable} that utilizes a decision tree to imitate the behavior of a given neural network teacher policy. \citet{bastani2018verifiable} utilizes the best DRL policy networks as its teacher policy. While VIPER lacks the ability to synthesize looping behaviors, it can be effectively employed for evaluating other approaches that utilize a program embedding space to synthesize more complex programs.

\subsection{LEAPS}
LEAPS is a programmatic RL framework introduced by \citet{trivedi2021learning}. The training framework of LEAPS consists of two stages. In the first stage, a model with an encoder-decoder architecture is trained to learn a continuous program embedding space. In the second stage, the Cross-Entropy Method \cite{rubinstein1997optimization} is utilized to search through the learned program embedding space and optimize the program policy for each task.

\subsection{LEAPS-ours}

LEAPS-ours utilizes the same framework as LEAPS but is trained on our proposed dataset while learning a program embedding space.

\section{Problem Set Details}
\label{sec:app_problem_set}

\subsection{\textsc{Karel} Problem Set Details}
\label{sec:app_karel}

The \textsc{Karel} problem set introduced in~\cite{trivedi2021learning} consists of six different tasks: \textsc{StairClimber},  \textsc{FourCorner}, \textsc{TopOff}, \textsc{Maze}, \textsc{CleanHouse} and \textsc{Harvester}. 
The performance of the policy networks is measured by averaging the rewards obtained from 10 randomly generated initial configurations of the environment.
All experiments are conducted on an $8\times8$ grid, except for the \textsc{CleanHouse} task.
\myfig{fig:karel both env} visualizes one of the random initial configurations and its ideal end state for each \textsc{Karel} task.

\input{figure/karel_example}

\myparagraph{\textsc{StairClimber}}
In this task, the agent is asked to move along the stair to reach the marked grid. The initial location of the agent and the marker is randomized near the stair, with the marker placed on the higher end. The reward is defined as $1$ if the agent successfully reaches the marked grid and $0$ otherwise.

\myparagraph{\textsc{FourCorner}}
The goal of the agent in this task is to place a marker on each corner of the grid to earn the reward. If any marker is placed in the wrong location, the reward is $0$. The initial position of the agent is randomized near the wall. The reward is calculated by multiplying the number of correctly placed markers by $0.25$.


\myparagraph{\textsc{TopOff}}
In this task, the agent is asked to place markers on marked grids and reach the rightmost grid of the bottom row. The reward is defined as the consecutive correct states of the last rows until the agent puts a marker on an empty location or does not place a marker on a marked grid. If the agent successfully reaches the rightmost grid of the last row, a bonus reward is granted. The agent is always initiated on the leftmost grid of the bottom row, facing east, while the locations of markers in the last row are randomized.


\myparagraph{\textsc{Maze}}
In this task, the agent has to navigate to reach the marked destination.  The locations of markers and the agent, as well as the configuration of the maze, are randomized. The reward is $1$ if the agent successfully reaches the marked grid or otherwise $0$.

\myparagraph{\textsc{CleanHouse}}
There is some garbage (markers) scattered around the apartment, and the agent is asked to clean them up.  The agent's objective is to collect as many markers as possible on the grid.  The apartment is represented by a 14x22 grid. While the agent's location remains fixed, the positions of the markers are randomized. The reward is calculated by dividing the number of markers collected by the total number of markers in the initial Karel state.

\myparagraph{\textsc{Harvester}}
The goal is to collect more markers on the grid, with markers appearing in all grids in the initial Karel environment. The reward is defined as the number of collected markers divided by the total markers in the initial state. 



\subsection{\textsc{Karel-Hard} Problem Set Details}
\label{sec:app_karel_hard}

Since all the tasks in the original Karel benchmark are well-solved by our method, we proposed a newly designed Karel-Hard benchmark to further evaluate the capability of HPRL. We define the state transition functions and reward functions for \textsc{DoorKey}, \textsc{OneStroke}, \textsc{Seeder}, and \textsc{Snake} based on Karel states. Each task includes more constraints and more complex structures, \eg two-phase structure for \textsc{DoorKey}, the restriction of no revisiting for \textsc{OneStroke}. 

The performance of the policy networks is measured by averaging the rewards of 10 random environment initial configurations.
The range of cumulative reward in all \textsc{Karel-Hard} tasks is [$0.0, 1.0$].
Figure \ref{fig:karel hard both env}  visualizes one of the random initial configurations and its ideal end state for each \textsc{Karel-Hard} task.

\input{figure/karel_hard_example}

\myparagraph{\textsc{DoorKey}}
An $8 \times 8$ grid is split into two areas: a $6 \times 3$ left chamber and a $6 \times 2$ right chamber. The two chambers are unconnected in the beginning. The agent has to pick up the marker in the left chamber to unlock the door, and then get into the right chamber to place the marker on the top of the target(marker). The initial location of the agent, the key(marker) in the left room and the target(marker) in the right room are randomly initialized. The reward is defined as $0.5$ for picking up the key and the remaining $0.5$ for placing the marker on the marked grid.

\myparagraph{\textsc{OneStroke}}
The goal is to make the agent traverse all grids without revisiting. The visited grids will become a wall and the episode will terminate if the agent hits the wall. The reward is defined as the number of grids visited divided by the total empty grids in the initial Karel environment. The initial location of the agent is randomized.

\myparagraph{\textsc{Seeder}}
The goal is to put markers on each grid in the Karel environment. The episode will end if markers are repeatedly placed. The reward is defined as the number of markers placed divided by the total number of empty grids in the initial Karel environment.

\myparagraph{\textsc{Snake}}
In this task, the agent acts as the head of the snake, and the goal is to eat (\ie pass through) as much food (markers) as possible without hitting its body. There is always exactly one marker existing in the environment until $20$ markers have been eaten. Once the agent passes a marker, the snake's body length will increase by $1$, and a new marker will appear at another position in the environment. The reward is defined as the number of markers eaten divided by $20$.

\section{Hyperparameters and Settings}
\label{sec:app_hyperparameters}

\subsection{LEAPS}

\input{table/app_param_LEAPS}

\subsection{LEAPS-ours}
\input{table/app_param_LEAPSours}
\subsection{\method\\}
\input{table/app_param_HPRL}


\section{The \textsc{Karel} Program Datasets Generation}
\label{sec:app_datasets}

\input{table/app_dataset}

The Karel program dataset used in this work includes $1$ million program sequences, with $85\%$ as the training dataset and $15\%$ as the evaluation dataset.
In addition to sequences of program tokens, the \textsc{Karel} program dataset also includes execution demonstrations (\eg state transition and action sequence) of each program in the dataset, which can be used for the \textit{latent behavior reconstruction loss} described in Section \ref{sec:method_stage1}.

To further improve the data quality, we added some heuristic rules while selecting data to filter out the programs with repetitive or offsetting behavior. 
The unwanted programs that we drop while collecting data are mainly determined by the following rules:
\begin{itemize}
    \item Contradictory Primitive Actions: \texttt{turnLeft} followed by \texttt{turnRight}, \texttt{pickMarker} followed by \texttt{putMarker}, or vice versa.
    \item Meaningless Programs: \texttt{end\_state == start\_state} after program execution
    \item Repetitive behaviors: a program which has the longest common subsequence of tokens longer than 9
\end{itemize}

We further analyze the distribution of the generated program sequences based on the control flow (e.g., \texttt{IF, IFELSE}) and loop command (e.g., \texttt{WHILE, REPEAT}). The statistical probabilities of programs containing control flow or loop commands are listed in Table \ref{table:dataset}. Results show that more than 40\% of the programs in the collected program sequences contain at least one of the control of loop commands, ensuring the diversity of the generated programs.


\input{text/app/app_program}
\input{text/app/app_program_hard}

\section{Evaluation Metric for Learning to Synthesize Out-of-distributional Programs}
\label{sec:app_recon_details}

\myparagraph{Measure the Performance} 
To measure the performance of programs synthesized by different methods,
we first collect and execute each target program,
yielding a target state sequence $\tau_{target} = [s_1, s_2, \dotsc, s_{T_{target}}]$.
Then, we reset the Karel environment to the initial state $ s_1$.
For our proposed framework, 
we synthesize a sequence of programs with the following procedure and optimize a program reconstruction reward to match the target program.
As described in~\mysecref{sec:method_stage2},
at each macro training time step $n, 1 \leq n \leq |H|$, we collect the state sequence $\tau_{\mathcal{P}} = [s^{\mathcal{P}}_1, s^{\mathcal{P}}_2, \dotsc, s^{\mathcal{P}}_{T_\mathcal{P}}]$ from the executing of task-solving program $\mathcal{P} = \langle \rho_i|i=1,..,n \rangle$, and calculate the program reconstruction reward $r^n = 1 - \mathcal{D}(\tau_{target}, \tau_{\mathcal{P}})$ where $\mathcal{D}$ is the normalized Levenshtein distance.
For executing the programs synthesized by LEAPS and LEAPS-ours, 
we simply start executing programs after
resetting the Karel environment to the initial state $ s_1$ and calculating the return.




%

\section{Synthesized Programs}
\label{sec:app_generated_program_samples}

In this section, we provide qualitative results (\ie synthesized programs)
of our proposed framework (HPRL-PPO), LEAPS, and LEAPS-ours.
The programs synthesized for the tasks in the \textsc{Karel} problem set are shown in \myfig{fig:karel_program_examples_stairClimber_topOff_cleanHouse} (\textsc{StairClimber}, \textsc{TopOff}, and \textsc{CleanHouse}),  \myfig{fig:karel_program_examples_fourCorner_maze} (\textsc{FourCorner}, \textsc{Maze}), and \myfig{fig:karel_program_examples_harvester} (\textsc{Harvester}).
The programs synthesized for the tasks in the \textsc{Karel-Hard} problem set are shown in \myfig{fig:karel_hard_program_examples_doorKey} (\textsc{DoorKey}), \myfig{fig:karel_hard_program_examples_oneStroke}(\textsc{OneStroke}), and \myfig{fig:karel_hard_program_examples_seeder_snake_app} (\textsc{Seeder} and \textsc{Snake}).


%% file: figure/karel_example.tex
\begin{figure*}[ht]
    \centering
    \begin{subfigure}[b]{0.45\textwidth}
    \centering
    \includegraphics[trim=25 16 25 55, clip,height=0.45\textwidth]{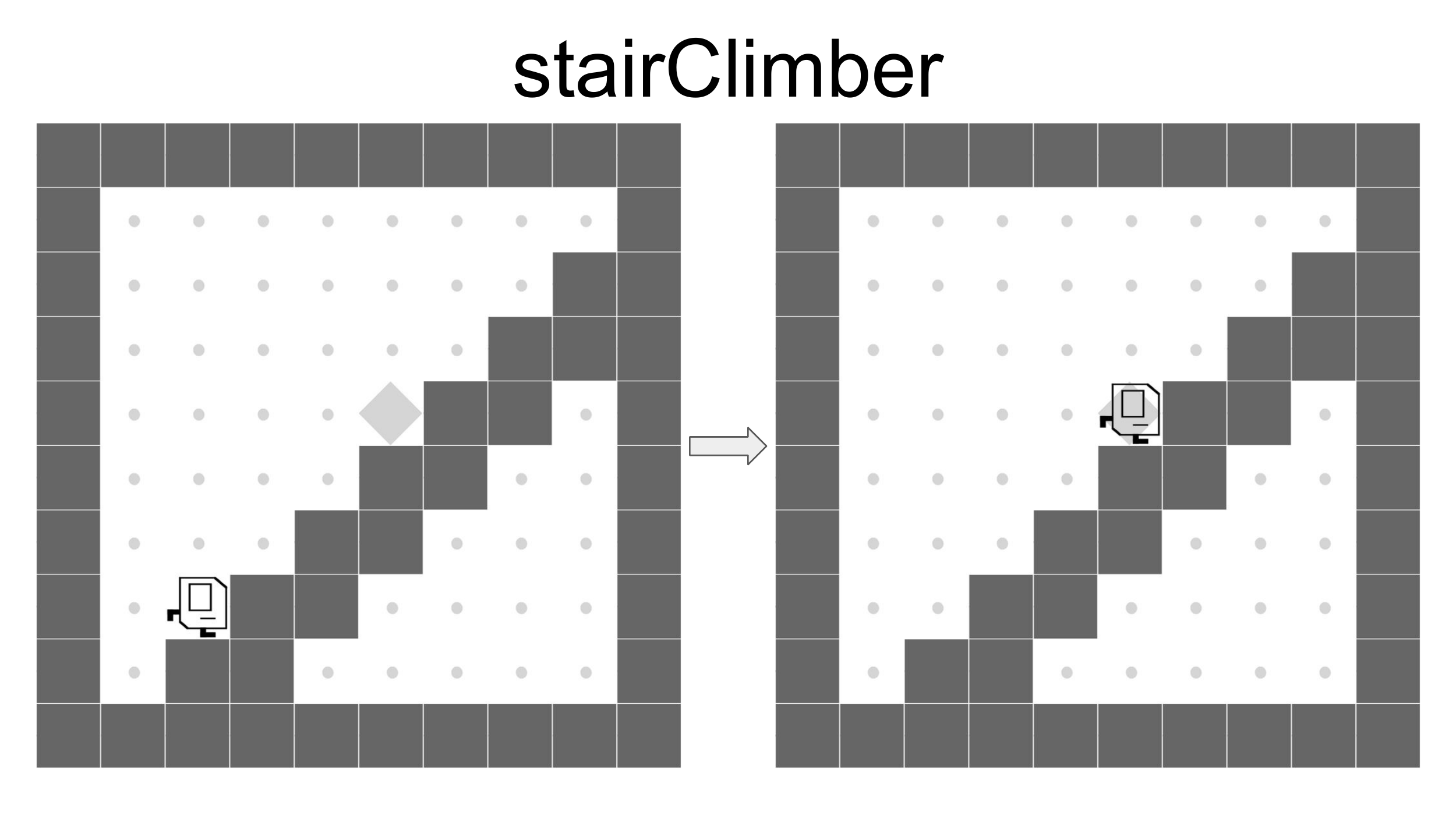}
    \caption{\textsc{StairClimber}}
    \end{subfigure}
    \hspace{0.5cm}
    \begin{subfigure}[b]{0.45\textwidth}
    \centering
    \includegraphics[trim=25 16 25 55, clip,height=0.45\textwidth]{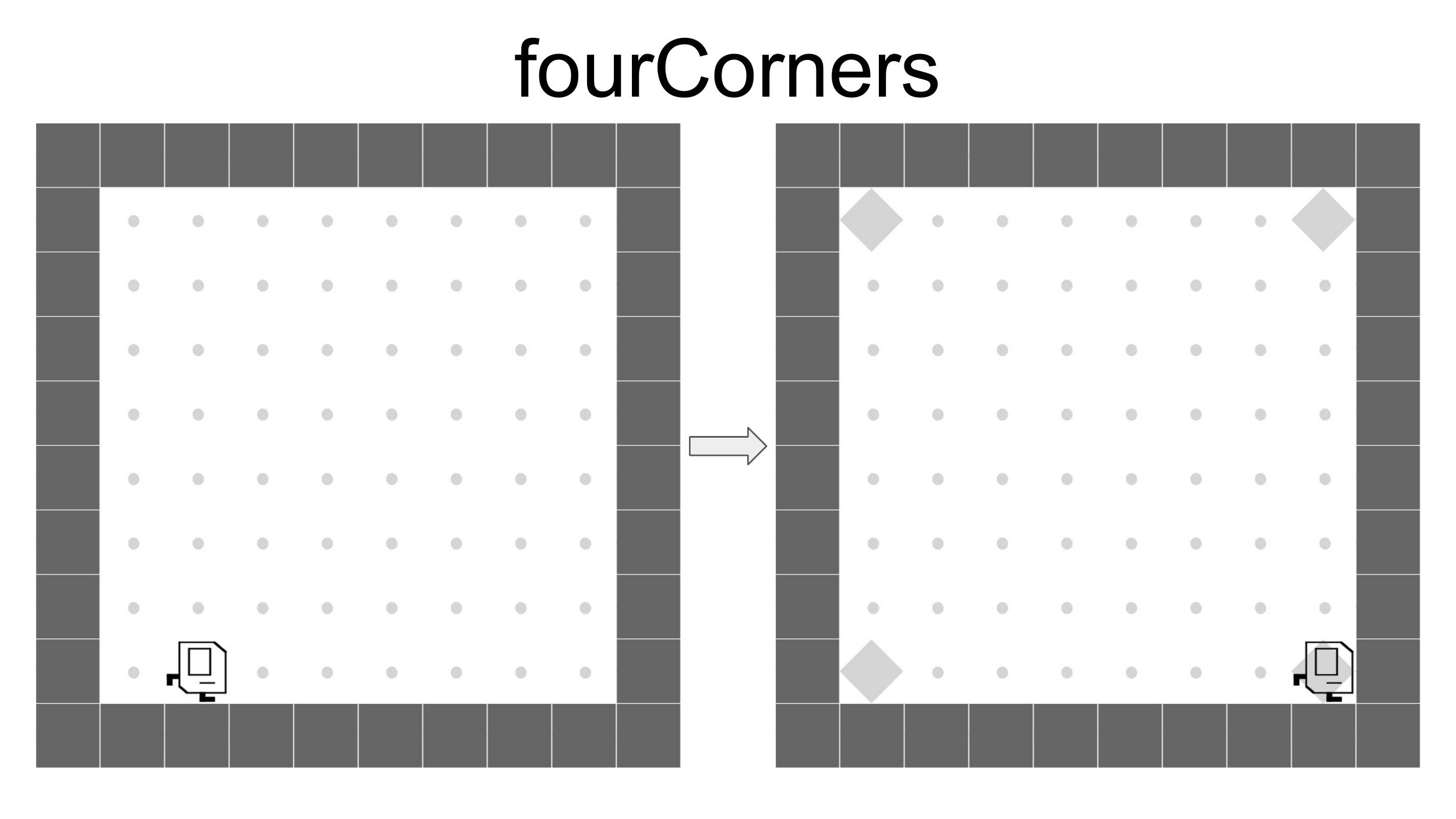}
    \caption{\textsc{fourCorner}}
    \end{subfigure}
    \hspace{0.5cm}
    \begin{subfigure}[b]{0.45\textwidth}
    \centering
    \includegraphics[trim=25 16 25 55, clip,height=0.45\textwidth]{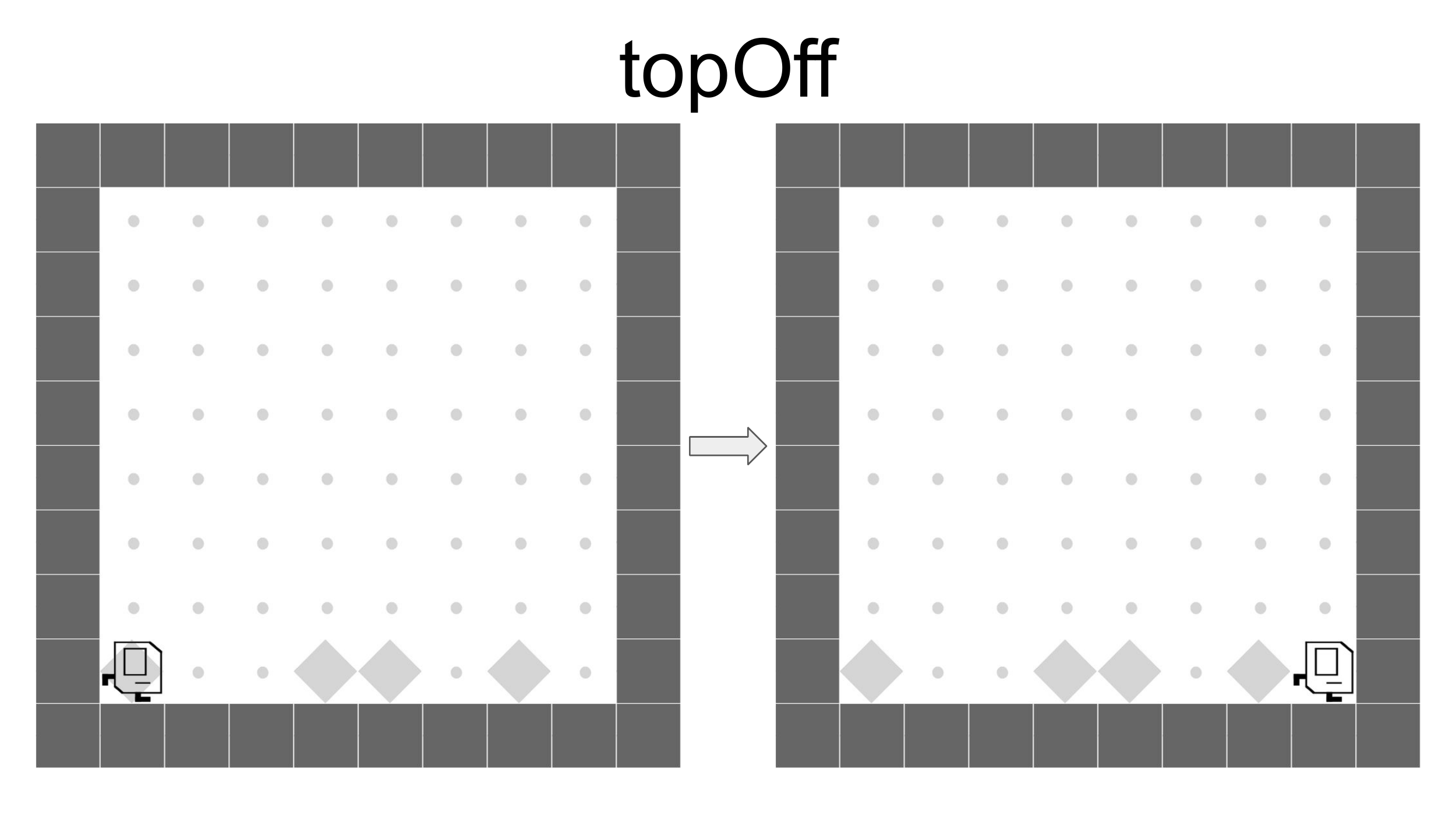}
    \caption{\textsc{TopOff}}
    \end{subfigure}
    \hspace{0.5cm}
    \begin{subfigure}[b]{0.45\textwidth}
    \centering
    \includegraphics[trim=25 16 25 50, clip,height=0.45\textwidth]{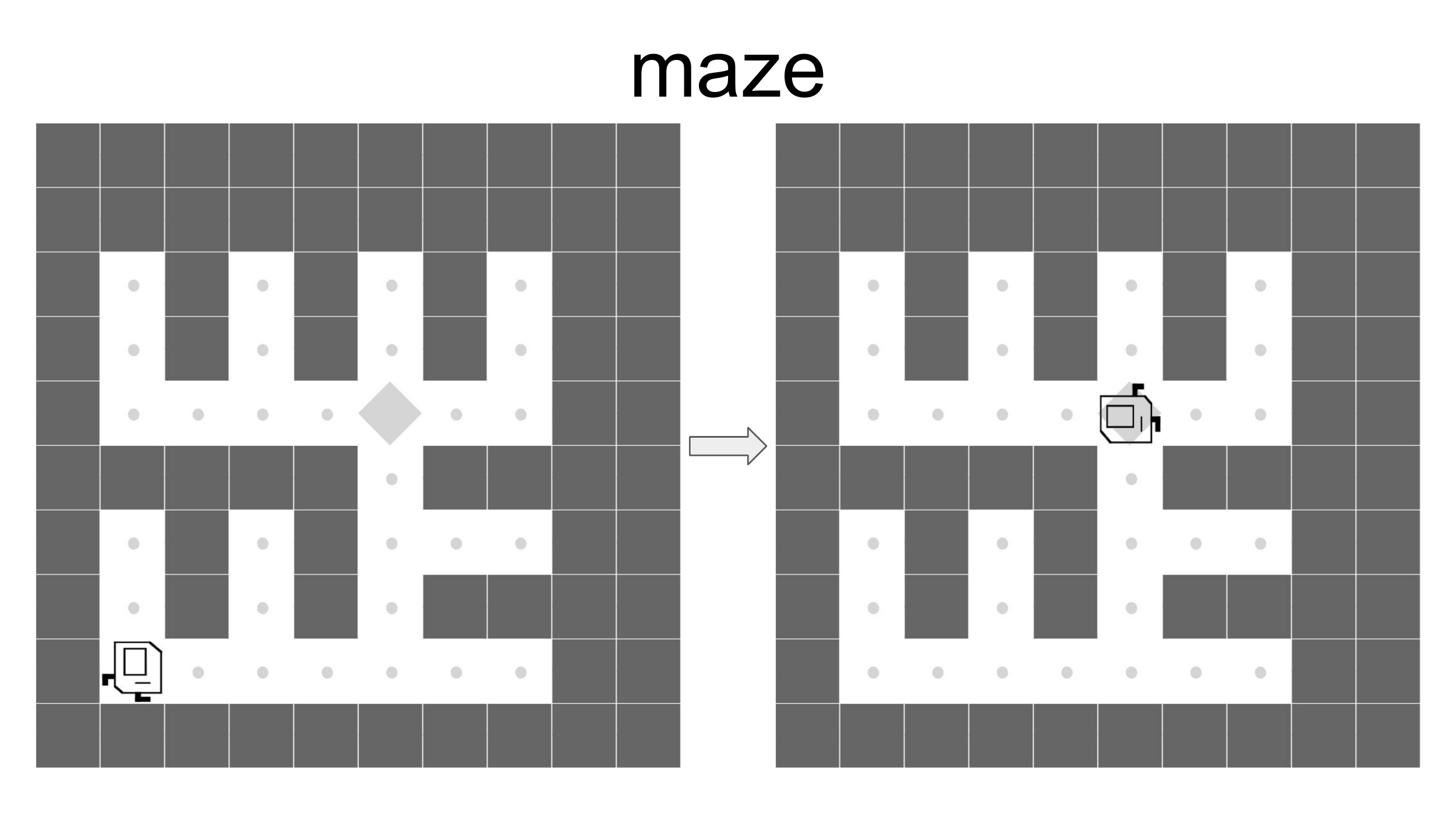}
    \caption{\textsc{Maze}}
    \end{subfigure}
    \hspace{0.5cm}
    \begin{subfigure}[b]{0.45\textwidth}
    \centering
    \includegraphics[trim=25 16 25 50, clip,height=0.45\textwidth]{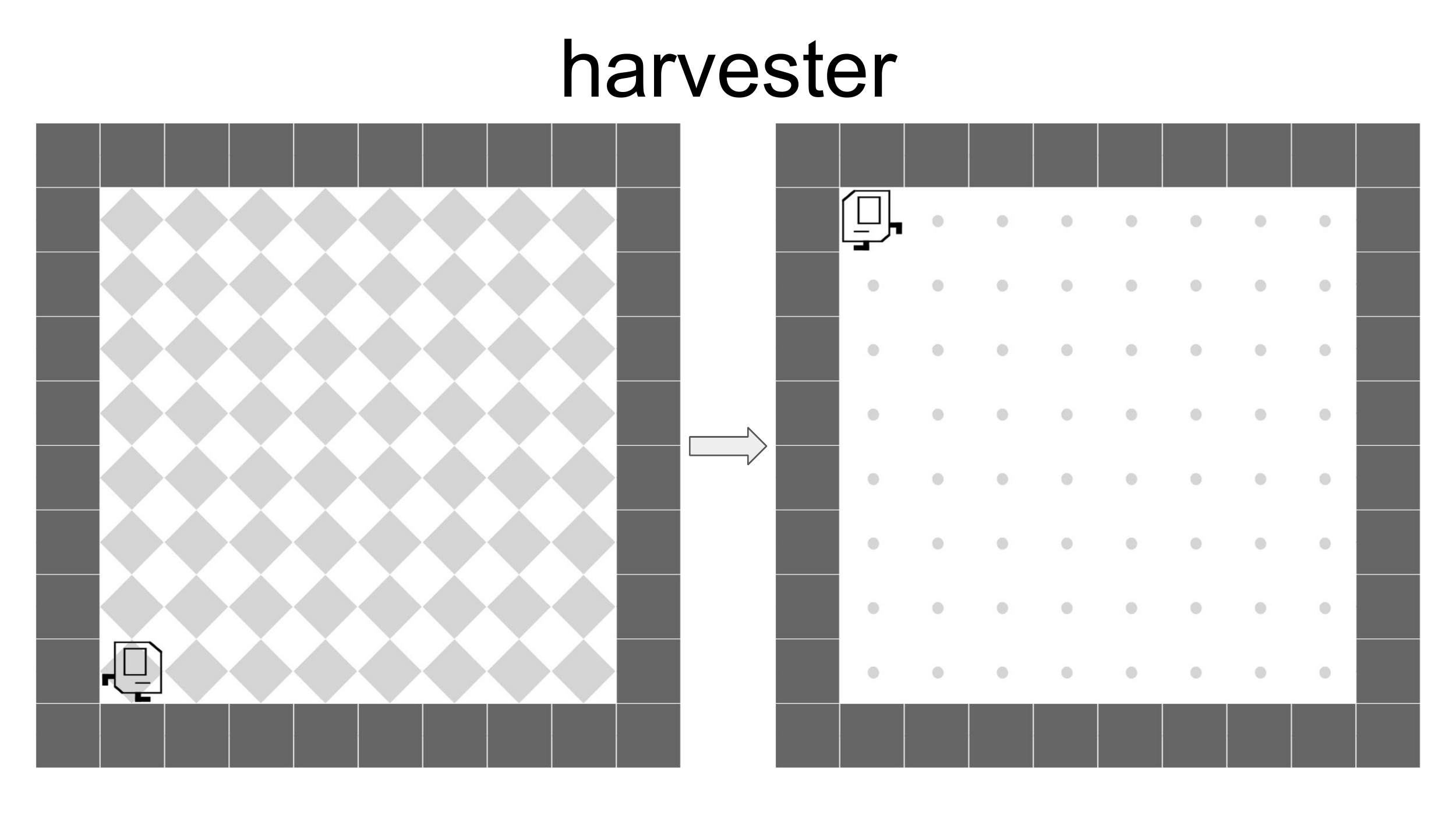}
    \caption{\textsc{Harvester}}
    \end{subfigure}
    \hspace{0.5cm}
    \begin{subfigure}[b]{0.9\textwidth}
    \centering
    \includegraphics[trim=25 0 25 30, clip,width=\textwidth]{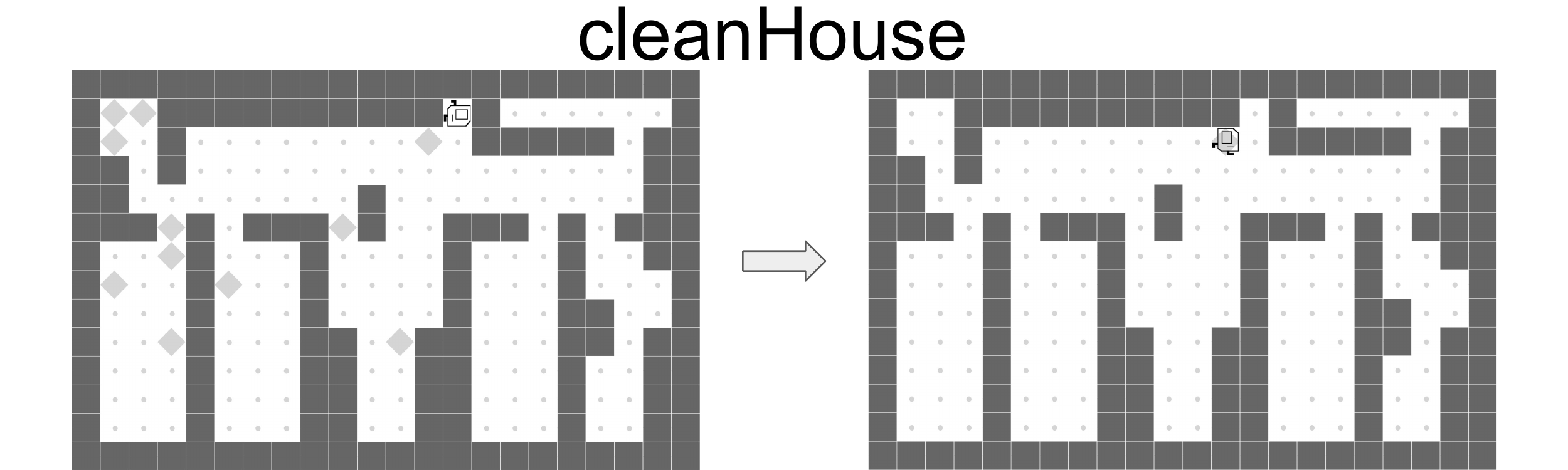}
    \caption{\textsc{CleanHouse}}
    \end{subfigure}
    \hspace{0.5cm}
    \caption[Karel Task Start/End State Depictions]{
    Illustrations of the initial and desired final state of each task in the \textsc{Karel} Problem set introduced in by \citet{trivedi2021learning}. Note that these illustrations are from \citep{trivedi2021learning}. The position of markers, walls, and agent's position are randomly set according to the configurations of each tasks. More details are provided in \mysecref{sec:app_karel}.
    }
    \label{fig:karel both env}
\end{figure*}

%% file: figure/karel_hard_example.tex
\begin{figure*}[ht]
    \centering
    \begin{subfigure}[b]{0.45\textwidth}
    \centering
    \includegraphics[trim=0 55 0 155, clip,height=0.45\textwidth]{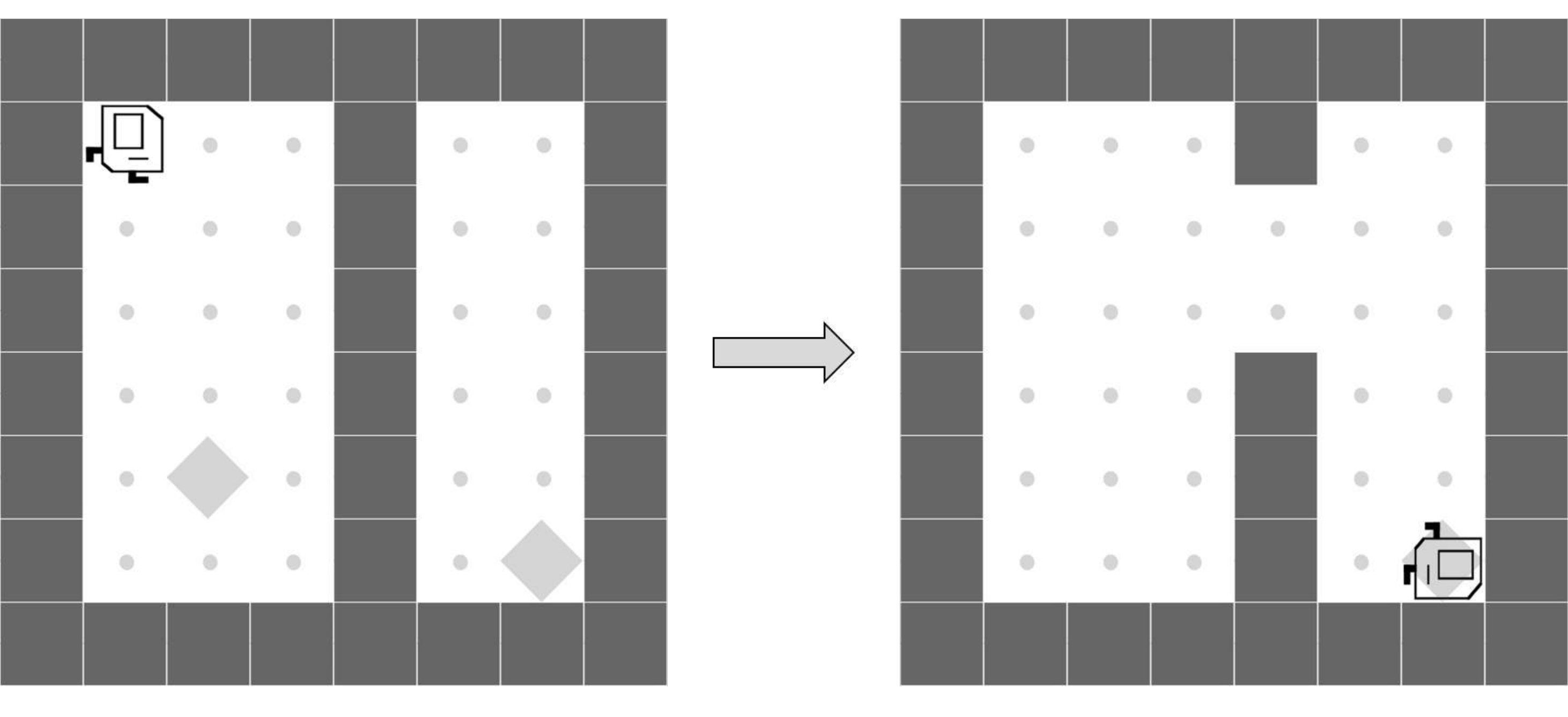}
    \caption{\textsc{DoorKey}}
    \end{subfigure}
    \hspace{0.5cm}
    \begin{subfigure}[b]{0.45\textwidth}
    \centering
    \includegraphics[trim=0 55 0 155, clip,height=0.45\textwidth]{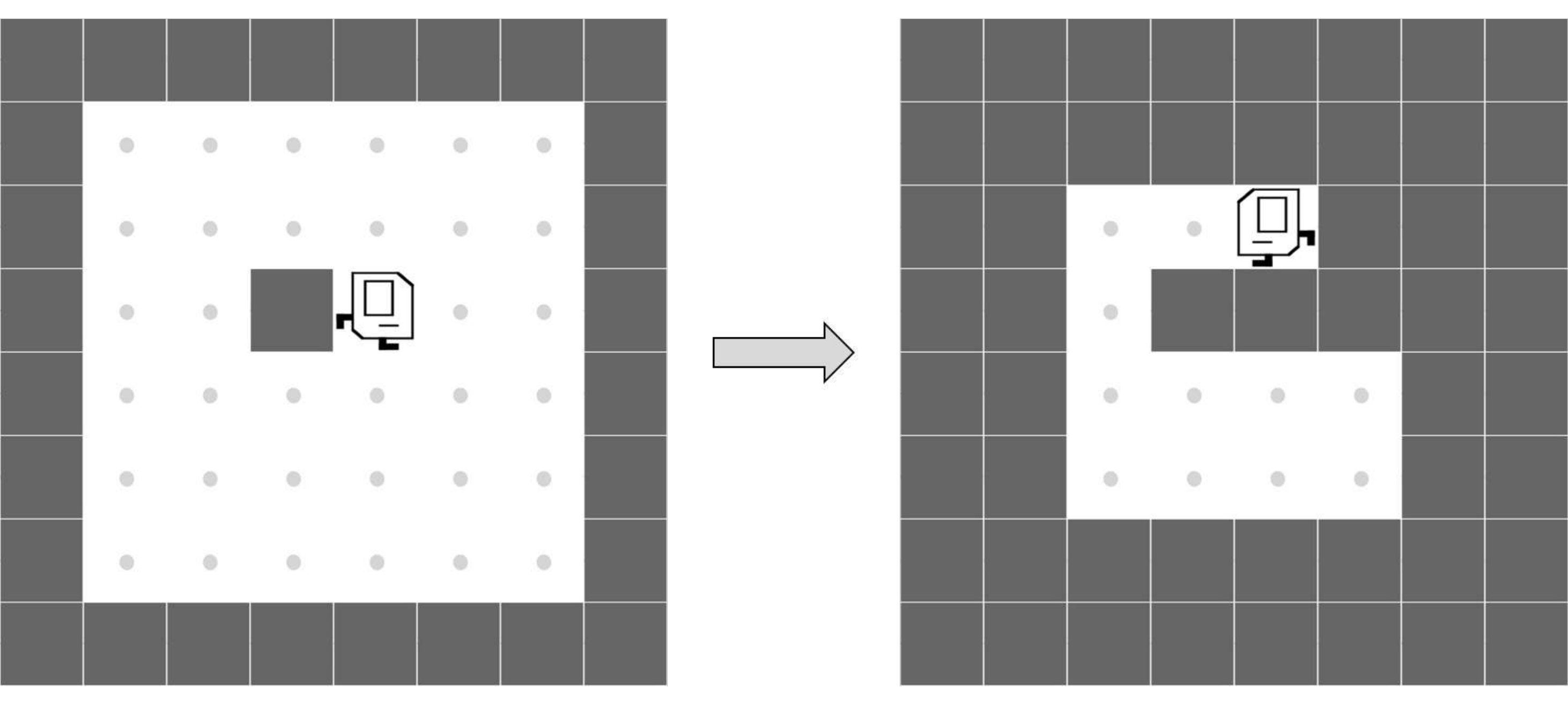}
    \caption{\textsc{OneStroke}}
    \end{subfigure}
    \hspace{0.5cm}
    
    \begin{subfigure}[b]{0.45\textwidth}
    \centering
    \includegraphics[trim=0 55 0 155, clip,height=0.45\textwidth]{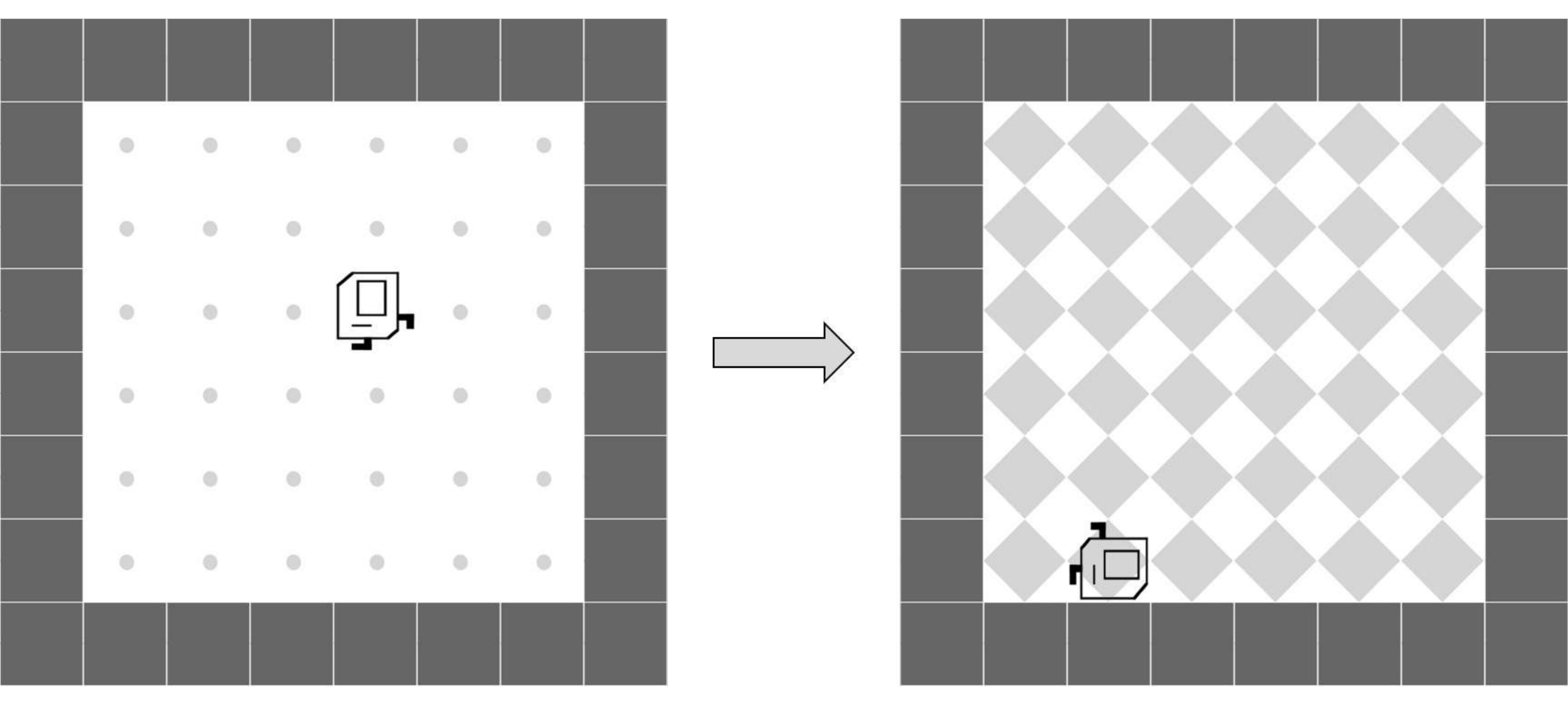}
    \caption{\textsc{Seeder}}
    \end{subfigure}
    \hspace{0.5cm}
    \begin{subfigure}[b]{0.45\textwidth}
    \centering
    \includegraphics[trim=0 55 0 155, clip,height=0.45\textwidth]{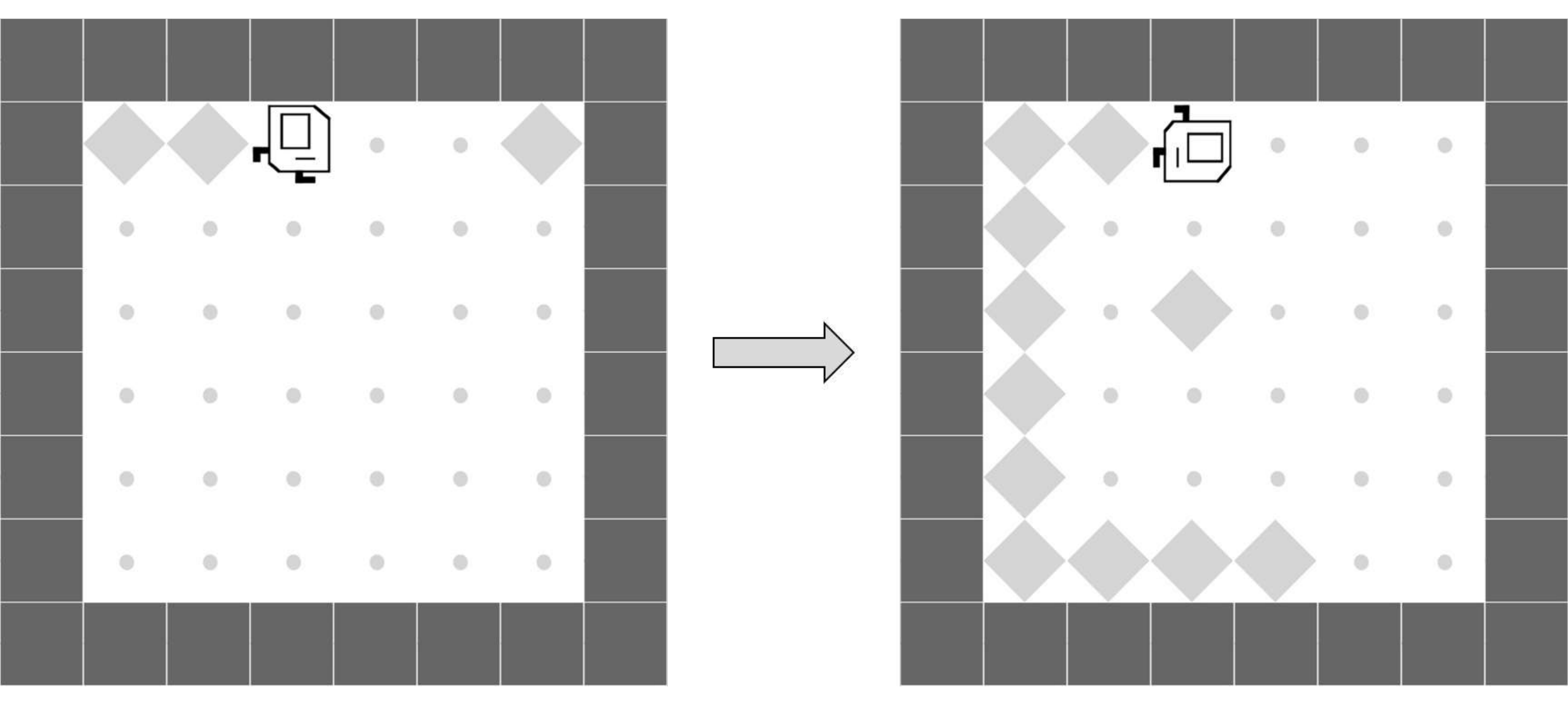}
    \caption{\textsc{Snake}}
    \end{subfigure}
    \hspace{0.5cm}
    \caption[Karel-Hard Task Start/End State Depictions]{
    Illustrations of the initial and final state of each task in the proposed \textsc{Karel-Hard} Problem Set. The position of markers, walls, and agent's position are randomly set according to the configurations of each tasks. More details are provided in \mysecref{sec:app_karel_hard}.
    }
    \label{fig:karel hard both env}
\end{figure*}

%% file: table/app_param_LEAPS.tex
Following the setting of LEAPS\cite{trivedi2021learning}, we experimented with sets of hyperparameters when searching the program embedding space to optimize the reward for both LEAPS and LEAPS-ours.
The LEAPS settings are described in~\mytable{table:karel_hard_setting} and~\mytable{table:recon_setting}.
S, $\sigma$, \# Elites, Exp Decay and $D_I$ represent population size, standard deviation, percentage of population elites, exponential $\sigma$ decay and initial distribution, respectively.

\textbf{Karel-Hard tasks}
\begin{table}[H]
\centering
\caption{LEAPS experiment settings on \textsc{Karel-Hard} tasks.}
\scalebox{0.9}{\begin{tabular}{@{}cccccc@{}}\toprule
LEAPS & S & $\sigma$ & $\#$ Elites & Exp Decay & $D_I$\\
\cmidrule{2-6}
\textsc{DoorKey} & 32 & 0.25 & 0.1 & False & $N(0,0.1I_d)$ \\
\textsc{OneStroke} & 64 & 0.5  & 0.05  & True & $N(1,0)$ \\
\textsc{Seeder} & 32 & 0.25  & 0.1  & False & $N(0,0.1I_d)$ \\
\textsc{Snake} & 32 & 0.25  & 0.2  & False & $N(0,I_d)$ \\
\bottomrule
\label{table:karel_hard_setting}
\end{tabular}}
\end{table}

\textbf{Reconstruction tasks}

\begin{table}[H]
\centering
\caption{LEAPS experiment settings on Program Reconstruction tasks.}
\scalebox{0.9}{\begin{tabular}{@{}cccccc@{}}\toprule
LEAPS & S & $\sigma$ & $\#$ Elites & Exp Decay & $D_I$\\
\cmidrule{2-6}
Len 25 & 32 & 0.5 & 0.05 & True & $N(0,I_d)$ \\
Len 50 & 32 & 0.5  & 0.2  & True & $N(0,0.1I_d)$ \\
Len 75 & 64 & 0.5  & 0.05  & True & $N(0,0.1I_d)$ \\
Len 100 & 64 & 0.5  & 0.1  & True & $N(0,0.1I_d)$ \\
\bottomrule
\label{table:recon_setting}
\end{tabular}}
\end{table}

%% file: table/app_param_LEAPSours.tex
\textbf{Karel tasks}

\begin{table}[H]
\centering
\caption[]{LEAPS-ours experiment settings on \textsc{Karel} tasks.}
\scalebox{0.9}{\begin{tabular}{@{}cccccc@{}}\toprule
LEAPS-ours & S & $\sigma$ & \# Elites & Exp Decay & $D_I$\\
\cmidrule{2-6}
\textsc{StairClimber} & 32 & 0.5 & 0.05 & True & $N(0,0.1I_d)$ \\
\textsc{FourCorners} & 32 & 0.5  & 0.1  & True & $N(1,0)$ \\
\textsc{TopOff} & 64 & 0.25  & 0.05  & True & $N(0,0.1I_d)$ \\
\textsc{Maze} & 64 & 0.1  & 0.2  & False & $N(1,0)$ \\
\textsc{CleanHouse} & 64 & 0.5  & 0.05  & True & $N(1,0)$ \\
\textsc{Harvester} & 64 & 0.5  & 0.05  & True & $N(1,0)$ \\
\bottomrule
\end{tabular}}
\end{table}

\textbf{Karel-Hard tasks}

\begin{table}[H]
\centering
\caption[]{LEAPS-ours experiment settings on \textsc{Karel-Hard} tasks.}
\scalebox{0.9}{\begin{tabular}{@{}cccccc@{}}\toprule
LEAPS-ours & S & $\sigma$ & \# Elites & Exp Decay & $D_I$\\
\cmidrule{2-6}
\textsc{DoorKey} & 64 & 0.5 & 0.2 & True & $N(1,0)$ \\
\textsc{OneStroke} & 64 & 0.5  & 0.05  & True & $N(1,0)$ \\
\textsc{Seeder} & 64 & 0.5  & 0.05  & True & $N(0,0.1, I_d)$ \\
\textsc{Snake} & 32 & 0.25  & 0.05  & False & $N(0,0.1, I_d)$ \\
\bottomrule
\end{tabular}
}
\end{table}

%% file: table/app_param_HPRL.tex

\textbf{Pretraining VAE} 

\begin{table}[H]
\centering
\label{tab:app_HPRL_hyperparameters}
\caption{Hyperparameters of VAE Pretraining}
\begin{tabular}{cc}
\hline
Parameter                        & Value/Setting \\ \hline
Latent Embedding Size            & 64            \\
GRU Hidden Layer Size            & 256           \\
\# GRU Layer for Encoder/Decoder & 1             \\
Batch Size                       & 256           \\
Nonlinearity                     & Tanh          \\
Learning Rate                    & 0.001         \\
Latent Loss Coefficient($\beta$) & 0.1           \\ \hline
\end{tabular}
\end{table}

\textbf{RL training on Meta Policies}


The Hyperparameters for HPRL-PPO and HPRL-SAC training are reported in Table \ref{tab:hyperparam_hprl_sac_ppo}.
For each task, we test on 5 different random seeds and take the average to measure the performance.





\begin{table} [h]
\centering
\caption{Hyperparameters of HPRL-PPO and HPRL-SAC Training}
\label{tab:hyperparam_hprl_sac_ppo}
\begin{tabular}{cll}
\hline
\begin{tabular}[c]{@{}c@{}}Training\\ Settings\end{tabular} & SAC  & PPO \\ \hline
Max \# Subprogram                                          & 5    & 5   \\
Max Subprogram Length                                      & 40   & 40  \\
Batch Size                                                 & 1024 & 256 \\ \hline
Specific Parameters &
  \begin{tabular}[c]{@{}l@{}}Init. Temperature: 0.0002\\ Actor Update Frequency: 200\\ Critic Target Update Frequency: 200\\ Num Seed Steps: 20000\\ Replay Buffer Size: 5M\\ Training Steps: 25M\\ Alpha Learning Rate: 0.0001\\ Actor Learning Rate: 0.0001\\ Critic Learning Rate: 0.00001\\ $\beta$: {[}0.9, 0.999{]}\\ Critic $\tau$: 0.005\\ Number of parallel actors: 16 \\ Discount factor: 0.99\\ Q-critic Hidden Dimension: 16\end{tabular} &
  \begin{tabular}[c]{@{}l@{}}Learning Rate: 0.00005\\ Entropy Coefficient: 0.1\\ Rollout Size: 12800\\ Eps: 0.00001\\ $\alpha$: 0.99\\ $\gamma$: 0.99\\ Use GAE: True\\ GAE lambda: 0.95\\ Value Loss Coefficient: 0.5\\ Clip Param: 0.2\\ Max grad. norm.: 0.5\\ Number of mini-batches: 10 \\ Update Epochs: 3\\ Training Steps: 25M\end{tabular} \\ \hline
\end{tabular}
\end{table}

%% file: table/app_dataset.tex
\begin{table*}
\centering
\caption[]{The statistical distribution of programs containing each token in our generated dataset.}
\scalebox{0.9}{\begin{tabular}{@{}ccccc@{}}\toprule
 & IFELSE & IF & WHILE & REPEAT \\
\cmidrule{2-5}
Our Dataset & 41\% & 47\% & 54\% & 22\% \\
\bottomrule
\end{tabular}}
\label{table:dataset}
\end{table*}

%% file: text/app/app_program.tex
\begin{figure*}[t]
\centering
\begin{mdframed}[frametitle=Karel Programs]
\begin{subfigure}[t]{0.90\textwidth}

\centering
\textbf{\textsc{StairClimber}}

{
\begin{subfigure}[t]{0.32\textwidth}
LEAPS
\begin{lstlisting}
DEF run m( 
    WHILE c( noMarkersPresent c) w( 
        turnRight 
        move 
        w) 
    WHILE c( rightIsClear c) w( 
        turnLeft 
        w) 
    m)
\end{lstlisting}
\end{subfigure}
\begin{subfigure}[t]{0.32\textwidth}
LEAPS-ours
\begin{lstlisting}
DEF run m( 
    turnRight 
    turnRight 
    WHILE c( noMarkersPresent c) w( 
        turnRight 
        move 
        w) 
    m)
\end{lstlisting}
\end{subfigure}
\begin{subfigure}[t]{0.32\textwidth}
HPRL-PPO
\begin{lstlisting}
DEF run m( 
    WHILE c( noMarkersPresent c) w( 
            turnRight 
            move 
            turnRight 
            move 
        w) 
    m)
\end{lstlisting}
\end{subfigure}
}
\end{subfigure}

\begin{subfigure}[t]{0.90\textwidth}

\centering
\textbf{\textsc{TopOff}}

{
\begin{subfigure}[t]{0.32\textwidth}
LEAPS
\begin{lstlisting}
DEF run m( 
    WHILE c( noMarkersPresent c) w( 
        move 
        w) 
    putMarker 
    move 
    WHILE c( not c( markersPresent c) c) w( 
        move 
        w)
    putMarker 
    move 
    WHILE c( not c( markersPresent c) c) w( 
        move 
        w) 
    putMarker 
    move 
    turnRight 
    turnRight 
    turnRight 
    turnRight 
    turnRight 
    turnRight
    turnRight 
    turnRight 
    m)
\end{lstlisting}
\end{subfigure}
\begin{subfigure}[t]{0.32\textwidth}
LEAPS-ours
\begin{lstlisting}
DEF run m( 
    WHILE c( not c( rightIsClear c) c) w( 
        WHILE c( not c( markersPresent c) c) w( 
            move 
            w) 
        putMarker 
        move 
        w) 
    WHILE c( not c( rightIsClear c) c) w( 
        pickMarker
        w) 
    m)
\end{lstlisting}
\end{subfigure}
\begin{subfigure}[t]{0.32\textwidth}
HPRL-PPO
\begin{lstlisting}
DEF run m( 
    REPEAT R=5 r( 
        move 
        WHILE c( noMarkersPresent c) w( 
            move 
            w) 
        putMarker 
        r)
    m)
\end{lstlisting}
\end{subfigure}
}
\end{subfigure}

\begin{subfigure}[t]{0.90\textwidth}

\centering
\textbf{\textsc{CleanHouse}}

{
\begin{subfigure}[t]{0.32\textwidth}
LEAPS
\begin{lstlisting}
DEF run m( 
    WHILE c( noMarkersPresent c) w( 
        turnRight 
        move 
        move 
        turnLeft 
        turnRight 
        pickMarker 
        w) 
    turnLeft 
    turnRight 
    m)
\end{lstlisting}
\end{subfigure}
\begin{subfigure}[t]{0.32\textwidth}
LEAPS-ours
\begin{lstlisting}
DEF run m( 
    move 
    WHILE c( noMarkersPresent c) w( 
        turnRight 
        move 
        WHILE c( frontIsClear c) w( 
            move 
            pickMarker 
            w) 
        w) 
    m) 
\end{lstlisting}
\end{subfigure}
\begin{subfigure}[t]{0.32\textwidth}
HPRL-PPO
\begin{lstlisting}
DEF run m( 
    WHILE c( noMarkersPresent c) w( 
        turnRight 
        move
        pickMarker 
        pickMarker 
        w) 
    m)
DEF run m( 
    WHILE c( noMarkersPresent c) w( 
        turnRight 
        move
        pickMarker 
        pickMarker 
        w) 
    m)
DEF run m( 
    WHILE c( noMarkersPresent c) w( 
        turnRight 
        move
        pickMarker 
        pickMarker 
        w) 
    m)
DEF run m( 
    WHILE c( noMarkersPresent c) w( 
        turnRight 
        move
        pickMarker 
        pickMarker 
        w) 
    m)
\end{lstlisting}
\end{subfigure}
}
\end{subfigure}
\end{mdframed}
\caption[\method\\ Karel Tasks Synthesized Programs]{\textbf{Example programs on Karel tasks: \textsc{StairClimber}, \textsc{TopOff} and \textsc{CleanHouse}.} The programs with best rewards out of all random seeds are shown.
}
\label{fig:karel_program_examples_stairClimber_topOff_cleanHouse}
\end{figure*}

\begin{figure*}[ht]
\begin{mdframed}
\begin{subfigure}[t]{0.90\textwidth}

\centering
\textbf{\textsc{FourCorner}}

{
\begin{subfigure}[t]{0.32\textwidth}
LEAPS
\begin{lstlisting}
DEF run m( 
    turnRight 
    move 
    turnRight 
    turnRight 
    turnRight 
    WHILE c( frontIsClear c) w( 
        move 
        w) 
    turnRight 
    putMarker
    WHILE c( frontIsClear c) w( 
        move 
        w) 
    turnRight 
    putMarker 
    WHILE c( frontIsClear c) w( 
        move 
        w) 
    turnRight 
    putMarker
    WHILE c( frontIsClear c) w( 
        move 
        w) 
    turnRight 
    putMarker 
    m)
\end{lstlisting}
\end{subfigure}
\begin{subfigure}[t]{0.32\textwidth}
LEAPS-ours
\begin{lstlisting}
DEF run m( 
    REPEAT R=5 r( 
        WHILE c( frontIsClear c) w( 
            move 
            w) 
        IFELSE c( not c( rightIsClear c) c) i( 
            turnLeft 
            putMarker 
            i) 
        ELSE e( 
            putMarker 
            e) 
        r)
    m)
\end{lstlisting}
\end{subfigure}
\begin{subfigure}[t]{0.32\textwidth}
HPRL-PPO
\begin{lstlisting}
DEF run m( 
    move 
    WHILE c( frontIsClear c) w( 
        move 
        w) 
    putMarker 
    turnLeft 
    m)
DEF run m( 
    move 
    WHILE c( frontIsClear c) w( 
        move 
        w) 
    putMarker 
    turnLeft 
    m)
DEF run m( 
    move 
    WHILE c( frontIsClear c) w( 
        move 
        w) 
    putMarker 
    turnLeft 
    m)
DEF run m( 
    move 
    WHILE c( frontIsClear c) w( 
        move 
        w) 
    putMarker 
    turnLeft 
    m)
\end{lstlisting}
\end{subfigure}
}
\end{subfigure}

\begin{subfigure}[t]{0.90\textwidth}

\centering
\textbf{\textsc{Maze}}

{
\begin{subfigure}[t]{0.32\textwidth}
LEAPS
\begin{lstlisting}
DEF run m( 
    IF c( frontIsClear c) i( 
        turnLeft 
        i) 
    WHILE c( noMarkersPresent c) w( 
        turnRight 
        move 
        w) 
    m)
\end{lstlisting}
\end{subfigure}
\begin{subfigure}[t]{0.32\textwidth}
LEAPS-ours
\begin{lstlisting}
DEF run m( 
    WHILE c( noMarkersPresent c) w( 
        turnRight 
        move 
        w) 
    turnRight 
    turnRight 
    turnRight
    m)
\end{lstlisting}
\end{subfigure}
\begin{subfigure}[t]{0.32\textwidth}
HPRL-PPO
\begin{lstlisting}
DEF run m( 
    WHILE c( noMarkersPresent c) w( 
        turnRight 
        move 
        w) 
    WHILE c( noMarkersPresent c) w( 
        turnRight 
        move 
        w) 
    m)
\end{lstlisting}
\end{subfigure}
}
\end{subfigure}
\end{mdframed}
\caption[\method\\ Karel Tasks Synthesized Programs]{\textbf{Example programs on Karel tasks: \textsc{FourCorner} and \textsc{Maze}.} The programs with best rewards out of all random seeds are shown.
}
\label{fig:karel_program_examples_fourCorner_maze}
\end{figure*}

\begin{figure*}[h]
\centering
\begin{mdframed}
\begin{subfigure}[t]{0.90\textwidth}

\centering
\textbf{\textsc{Harvester}}

{
\begin{subfigure}[t]{0.32\textwidth}
LEAPS
\begin{lstlisting}
DEF run m( 
    turnLeft 
    turnLeft 
    pickMarker 
    move 
    pickMarker 
    pickMarker 
    move 
    pickMarker 
    move 
    pickMarker 
    move 
    pickMarker 
    move 
    turnLeft
    pickMarker 
    move 
    pickMarker 
    move 
    pickMarker 
    move 
    pickMarker 
    move 
    turnLeft 
    pickMarker 
    move 
    pickMarker 
    move 
    pickMarker 
    move
    pickMarker 
    move 
    turnLeft 
    pickMarker 
    move 
    pickMarker 
    move 
    pickMarker 
    move 
    m)
\end{lstlisting}
\end{subfigure}
\begin{subfigure}[t]{0.32\textwidth}
LEAPS-ours
\begin{lstlisting}
DEF run m( 
    WHILE c( leftIsClear c) w( 
        REPEAT R=4 r( 
            pickMarker 
            move 
            r) 
        turnLeft 
        pickMarker 
        move 
        turnLeft 
        pickMarker 
        move 
        w) 
        turnLeft 
        pickMarker 
        turnLeft
    m)
\end{lstlisting}
\end{subfigure}
\begin{subfigure}[t]{0.32\textwidth}
HPRL-PPO
\begin{lstlisting}
DEF run m( 
    REPEAT R=4 r( 
        REPEAT R=4 r( 
            pickMarker 
            turnRight 
            move 
            pickMarker 
            turnRight 
            move 
            pickMarker 
            move 
            pickMarker 
            move 
            r) 
        turnRight 
        pickMarker 
        move 
        pickMarker 
        move 
        pickMarker 
        move 
        r)
    m)
\end{lstlisting}
\end{subfigure}
}
\end{subfigure}
\end{mdframed}
\caption[\method\\ Karel Tasks Synthesized Programs]{\textbf{Example programs on Karel tasks: \textsc{Harvester}.} The programs with best rewards out of all random seeds are shown.
}
\label{fig:karel_program_examples_harvester}
\end{figure*}

%% file: text/app/app_program_hard.tex
\begin{figure*}[h]
\centering
\begin{mdframed}[frametitle=Karel-Hard Programs]
\begin{subfigure}[t]{0.90\textwidth}

\centering
\textbf{\textsc{DoorKey}}

{
\begin{subfigure}[t]{0.32\textwidth}
LEAPS
\begin{lstlisting}
DEF run m( 
    move 
    turnRight 
    putMarker 
    pickMarker 
    move 
    WHILE c( leftIsClear c) w( 
        pickMarker 
        move 
        w) 
    m) 
\end{lstlisting}
\end{subfigure}
\begin{subfigure}[t]{0.32\textwidth}
LEAPS-ours
\begin{lstlisting}
DEF run m( 
    WHILE c( rightIsClear c) w( 
        turnRight 
        pickMarker 
        turnLeft 
        pickMarker 
        pickMarker 
        pickMarker 
        pickMarker
        move 
        turnLeft 
        move 
        w) 
    m) 
\end{lstlisting}
\end{subfigure}
\begin{subfigure}[t]{0.32\textwidth}
HPRL-PPO
\begin{lstlisting}
DEF run m( 
    REPEAT R=4 r( 
        REPEAT R=4 r( turnRight move pickMarker move pickMarker move r) 
        pickMarker move r) 
    m)
DEF run m( 
    REPEAT R=5 r( 
        turnRight move 
        REPEAT R=5 r( move r) 
        move pickMarker move 
        r) 
    m)
DEF run m( 
    REPEAT R=4 r( 
        REPEAT R=4 r( 
            turnRight move 
            REPEAT R=3 r( 
                move pickMarker 
                move pickMarker r) 
            r) 
        r) 
    m)
DEF run m( 
    REPEAT R=4 r( 
        REPEAT R=4 r( 
            turnRight move 
            pickMarker move 
            pickMarker 
            REPEAT R=2 r( 
                pickMarker move 
                pickMarker pickMarker r) 
            r) 
        r) 
    m)
DEF run m( 
    REPEAT R=4 r( 
        turnRight 
        REPEAT R=4 r( 
            turnRight move move 
            pickMarker move 
            r) 
        move pickMarker move 
        r) 
    move pickMarker 
    m)
\end{lstlisting}
\end{subfigure}
}
\end{subfigure}
\end{mdframed}
\caption[\method\\ Karel Tasks Synthesized Programs]{\textbf{Example programs on Karel-Hard tasks: \textsc{DoorKey}.} The programs with best rewards out of all random seeds are shown.}
\label{fig:karel_hard_program_examples_doorKey}
\end{figure*}

\begin{figure*}[h]
\centering
\begin{mdframed}
\begin{subfigure}[t]{0.90\textwidth}

\centering
\textbf{\textsc{OneStroke}}

{
\begin{subfigure}[t]{0.32\textwidth}
LEAPS
\begin{lstlisting}
DEF run m( 
    REPEAT R=9 r( 
        turnRight 
        turnRight 
        WHILE c( frontIsClear c) w( 
            move 
            w) 
        turnRight 
        WHILE c( frontIsClear c) w( 
            move 
            w) 
        r) 
    turnRight 
    m)
\end{lstlisting}
\end{subfigure}
\begin{subfigure}[t]{0.32\textwidth}
LEAPS-ours
\begin{lstlisting}
DEF run m( 
    turnRight 
    WHILE c( frontIsClear c) w( 
        WHILE c( frontIsClear c) w( 
            WHILE c( frontIsClear c) w( 
                WHILE c( frontIsClear c) w( 
                    move 
                    w) 
                turnRight 
                w) 
            turnRight 
            w) 
        turnRight 
        w) 
    turnRight 
    m) 
\end{lstlisting}
\end{subfigure}
\begin{subfigure}[t]{0.32\textwidth}
HPRL-PPO
\begin{lstlisting}
DEF run m( 
    WHILE c( frontIsClear c) w( move w) turnRight 
    WHILE c( frontIsClear c) w( move w) turnRight 
    WHILE c( frontIsClear c) w( move w) turnRight 
    m)
DEF run m( 
    WHILE c( frontIsClear c) w( move w) turnRight 
    WHILE c( frontIsClear c) w( move w) turnRight 
    WHILE c( frontIsClear c) w( move w) turnRight 
    m)
DEF run m( 
    WHILE c( frontIsClear c) w( move w) turnRight 
    WHILE c( frontIsClear c) w( move w) turnRight 
    WHILE c( frontIsClear c) w( move w) turnRight 
    WHILE c( frontIsClear c) w( move w) turnRight 
    m)
DEF run m( 
    WHILE c( frontIsClear c) w( move w) turnRight 
    WHILE c( frontIsClear c) w( move w) turnRight 
    WHILE c( frontIsClear c) w( move w) turnRight 
    m)
\end{lstlisting}
\end{subfigure}
}
\end{subfigure}
\end{mdframed}
\caption[\method\\ Karel Tasks Synthesized Programs]{\textbf{Example programs on Karel-Hard tasks: \textsc{OneStroke}.} The programs with best rewards out of all random seeds are shown.}
\label{fig:karel_hard_program_examples_oneStroke}
\end{figure*}

\begin{figure*}[h]
\begin{mdframed}
\begin{subfigure}[t]{0.90\textwidth}

\centering
\textbf{\textsc{Seeder}}

{
\begin{subfigure}[t]{0.32\textwidth}
LEAPS
\begin{lstlisting}
DEF run m( 
    WHILE c( noMarkersPresent c) w( 
        turnRight 
        putMarker 
        move 
        move 
        w) 
    turnRight 
    turnRight 
    turnRight 
    turnRight 
    turnRight 
    turnRight 
    turnRight 
    turnRight 
    m) 
\end{lstlisting}
\end{subfigure}
\begin{subfigure}[t]{0.32\textwidth}
LEAPS-ours
\begin{lstlisting}
DEF run m( 
    WHILE c( noMarkersPresent c) w( 
        putMarker 
        move 
        turnRight 
        move 
        w) 
    turnRight 
    turnRight 
    turnRight 
    turnRight 
    turnRight 
    turnRight 
    turnRight 
    turnRight
    m)
\end{lstlisting}
\end{subfigure}
\begin{subfigure}[t]{0.32\textwidth}
HPRL-PPO
\begin{lstlisting}
DEF run m( 
    putMarker move 
    putMarker move 
    putMarker move 
    putMarker move 
    putMarker move 
    turnRight move 
    m)
DEF run m( 
    putMarker move 
    putMarker move 
    putMarker move 
    putMarker move 
    putMarker move 
    turnRight move 
    putMarker move 
    m)
DEF run m( 
    putMarker move 
    putMarker move 
    putMarker move 
    putMarker move 
    turnRight move 
    putMarker move 
    turnRight move 
    m)
DEF run m( 
    putMarker move 
    putMarker move 
    putMarker move 
    putMarker move 
    turnRight move 
    putMarker move 
    turnRight move
DEF run m( 
    putMarker move 
    putMarker move 
    putMarker move 
    putMarker move 
    turnRight move 
    putMarker move 
    turnRight move 
    m)
\end{lstlisting}
\end{subfigure}
}
\end{subfigure}

\begin{subfigure}[t]{0.90\textwidth}

\centering
\textbf{\textsc{Snake}}

{
\begin{subfigure}[t]{0.32\textwidth}
LEAPS
\begin{lstlisting}
DEF run m( 
    turnRight 
    turnLeft 
    pickMarker 
    move 
    move 
    move 
    WHILE c( rightIsClear c) w( 
        turnLeft 
        move 
        move 
        w) 
    turnLeft 
    turnLeft 
    turnLeft 
    turnLeft 
    m) 
\end{lstlisting}
\end{subfigure}
\begin{subfigure}[t]{0.32\textwidth}
LEAPS-ours
\begin{lstlisting}
DEF run m( 
    move 
    turnRight 
    pickMarker 
    pickMarker 
    WHILE c( rightIsClear c) w( 
        turnLeft 
        move 
        move 
        w) 
    turnRight 
    move 
    move 
    move 
    m)
\end{lstlisting}
\end{subfigure}
\begin{subfigure}[t]{0.32\textwidth}
HPRL-PPO
\begin{lstlisting}
DEF run m( 
    move 
    WHILE c( noMarkersPresent c) w( 
        move 
        move 
        turnLeft 
        w) 
    move 
    turnLeft 
    m)
DEF run m( 
    move 
    WHILE c( noMarkersPresent c) w( 
        move 
        move 
        turnLeft 
        w) 
    m)
DEF run m( 
    move 
    WHILE c( noMarkersPresent c) w( 
        move 
        move 
        turnLeft 
        w) 
    move 
    turnLeft 
    m)
\end{lstlisting}
\end{subfigure}
}
\end{subfigure}
\end{mdframed}
\caption[\method\\ Karel Tasks Synthesized Programs]{\textbf{Example programs on Karel-Hard tasks: \textsc{Seeder} and \textsc{Snake}.} The programs with best rewards out of all random seeds are shown.}
\label{fig:karel_hard_program_examples_seeder_snake_app}
\end{figure*}